\documentclass[10pt,twocolumn,letterpaper]{article}

\usepackage[pagenumbers]{wacv} 

\usepackage{graphicx}
\usepackage{amsmath}
\usepackage{amssymb}
\usepackage{booktabs}

\usepackage[pagebackref,breaklinks,colorlinks]{hyperref}
\usepackage{xspace}
\newcommand{\coolname}{GeoDiffuser\xspace}

\newenvironment{packed_itemize}
{\begin{itemize}
    \setlength{\itemsep}{1pt}
    \setlength{\parskip}{0pt}
    \setlength{\parsep}{0pt}
}{\end{itemize}}

\newcommand{\parahead}[1]{\noindent\textbf{#1}:~}

\newcommand{\manu}{manu.\xspace}

\usepackage{algorithm}
\usepackage{algpseudocode}
\usepackage{blindtext}
\usepackage{float}
\usepackage{lipsum,multicol}

\usepackage{xcolor}

\definecolor{cb-black}      {RGB}{  0,   0,   0}
\definecolor{cb-blue-green} {RGB}{  0,  073,  073}
\definecolor{cb-green-sea}  {RGB}{  0, 146, 146}
\definecolor{cb-rose}       {RGB}{255, 109, 182}
\definecolor{cb-salmon-pink}{RGB}{255, 182, 119}
\definecolor{cb-purple}     {RGB}{ 194,   106, 119}
\definecolor{cb-blue}       {RGB}{ 0, 109, 219}
\definecolor{cb-lilac}      {RGB}{182, 109, 255}
\definecolor{cb-blue-sky}   {RGB}{109, 182, 255}
\definecolor{cb-blue-light} {RGB}{182, 219, 255}
\definecolor{cb-burgundy}   {RGB}{146,   0,   0}
\definecolor{cb-brown}      {RGB}{146,  73,   0}
\definecolor{cb-clay}       {RGB}{219, 209,   0}
\definecolor{cb-green-lime} {RGB}{ 36, 255,  36}
\definecolor{cb-yellow}     {RGB}{255, 255, 109}

\definecolor{cb-edit}     {RGB}{ 60, 170, 200}

\newcommand{\cbEdit}  [1]{#1}
\usepackage[title]{appendix}

\usepackage[capitalize]{cleveref}
\crefname{section}{Sec.}{Secs.}
\Crefname{section}{Section}{Sections}
\Crefname{table}{Table}{Tables}
\crefname{table}{Tab.}{Tabs.}

\def\wacvPaperID{304} 
\def\confName{WACV}
\def\confYear{2025}

\begin{document}

\title{\coolname: Geometry-Based Image Editing with Diffusion Models
\vspace{-0.3cm}
}

\author{
Rahul Sajnani$^{1,2}$\qquad Jeroen Vanbaar$^{2}$\qquad Jie Min$^{2}$ \qquad Kapil Katyal$^{2}$ \qquad Srinath Sridhar$^{1,2}$\\ \vspace{1mm}
\text{\normalsize $^1$Brown University\qquad $^2$Amazon Robotics}\\
\href{https://ivl.cs.brown.edu/research/geodiffuser.html}{ivl.cs.brown.edu/research/geodiffuser}
}

\twocolumn[{%
\renewcommand\twocolumn[1][]{#1}%
\maketitle
\begin{center}
    \vspace{-1.cm}
    \centering
    \captionsetup{type=figure}
    \includegraphics[width=1.\textwidth]{images/teaser/Teaser_5.pdf}
    \vspace{-0.35in}
    \captionof{figure}{
    We introduce \textbf{\coolname}, a unified method to perform common 2D and 3D image editing tasks like object translation, 3D rotation, object removal, and re-scaling while preserving object style and inpainting disoccluded regions.
    Our method is a zero-shot optimization-based method that builds on top of a pre-trained diffusion model.
    We treat image editing as a geometric transformation of parts of the image and bake this directly into a shared attention-based edit optimization.
  In this figure, the top row shows natural images and the bottom row shows the edit.
    \vspace{-0.5cm}
    }
    \label{fig:teaser}
\end{center}%
}]

\vspace*{0.5em}
\begin{abstract}
The success of image generative models has enabled us to build methods that can edit images based on text or other user input.
However, these methods are imprecise, require additional information, or are limited to only 2D image edits.
We present \coolname, a zero-shot optimization-based method that unifies common 2D and 3D image-based object editing capabilities into a single method.
Our key insight is to view image editing operations as geometric transformations.
We show that these transformations can be directly incorporated into the attention layers in diffusion models to implicitly perform editing operations.
Our training-free optimization method uses an objective function that seeks to preserve object style but generate plausible images, for instance with accurate lighting and shadows.
It also inpaints disoccluded parts of the image where the object was originally located.
%
Given a natural image and user input, we segment the foreground object~\cite{kirillov2023segment} and estimate a corresponding transform which is used by our optimization approach for editing.
\Cref{fig:teaser} shows that \coolname can perform common 2D and 3D edits like object translation, 3D rotation, and removal.
We present quantitative results, including a perceptual study, that shows how our approach is better than existing methods.
%
\end{abstract}
\section{Introduction}
\label{sec:intro}
Image generative models have seen significant progress recently.
The most advanced diffusion-based models can now generate high-quality images almost indistinguishable from reality~\cite{ramesh2022hierarchical,Rombach_2022_CVPR,saharia2022photorealistic,yu2023scaling}.
These models generate images with the desired content and detail by conditioning on text prompts, sometimes in combination with additional information like segmentation masks~\cite{zhang2023adding}.
They have proliferated in use with many commercial products incorporating them~\cite{dalle,google_gemini,amazon_titan}.

Although realistic image generation is an important capability, in many cases, we may also want to edit generated or existing natural images.
While past work relied on computer graphics techniques for image editing~\cite{kholgade20143d,zheng2012interactive,chen20133,lalonde2007photo}, recent works have put generative models to use for this problem.
In particular, generative models have been shown to enable text-based edits~\cite{hertz2022prompt,mokady2023null,vinker2023concept}, object stitching~\cite{song2023objectstitch,dwibedi2017cut}, object removal~\cite{Rombach_2022_CVPR}, and interactive edits using user-defined points~\cite{pan2023drag,shi2023dragdiffusion,mou2023dragondiffusion}, 3D transforms~\cite{pandey2023diffusion} or flow~\cite{motion_guidance2024}.
However, these methods have important limitations.
Text-based editing methods are imprecise for edits requiring spatial control.
Object stitching and removal methods cannot easily be extended to geometric edits.
Finally, interactive point-/flow-based methods require additional input such as a text prompt or optical flow.

In this paper, we present \textbf{\coolname}, a method that unifies various image-based object editing capabilities into a single method.
We take the view that common user-specified image editing operations can be cast as \textbf{geometric transformations} of parts of the image.
For instance, 2D object translation or 3D object rotation can be represented as a bijective transformation of the foreground object.
%
%
However, naively applying this transformation on the image is unlikely to produce plausible edits, for instance, due to mismatched lighting or shadows.
To overcome this problem, we use diffusion models, specifically the general editing approach (see \Cref{fig:background}) enabled by DDIM Inversion~\cite{mokady2022nulltext}.
\cbEdit{Our key contribution is to \textbf{bake in the geometric transformation directly within the shared attention layers of a diffusion model to preserve style} while enabling a wide range of user-specified 2D and 3D edits.}
%
Additionally, \coolname is a zero-shot optimization-based method that operates \textbf{without the need for any additional training} and can support any diffusion model with attention layers.

\Cref{fig:teaser} shows common image edits performed by \coolname on natural images.
Without any hyperparameter tuning, our method can perform 2D edits like object translation or removal, or 3D edits like 3D rotation and translation.
Given a natural image, we first segment the object of interest~\cite{kirillov2023segment}, and optionally, extract a depth map~\cite{yang2024depth} for 3D edits.
For each type of edit, we first compute a geometric transform based on user input and formulate an objective function for optimization.
Unlike approaches that first `lift' an object from an image
and then stitch the transformed object back into the image~\cite{kholgade20143d}, we implicitly perform these steps by applying the transform directly to the self- and cross-attention layers.
Since attention captures both local and global image interactions, our results exhibit accurate lighting, shadows and reflection while inpainting the disoccluded image regions.
Moreover, our objective function incorporates terms to preserve the original style of the transformed object.
%
%
%

We show extensive qualitative results that demonstrate that our method can perform multiple 2D and 3D editing operations using a single approach.
To evaluate our method quantitatively, we provide experiments through a perceptual study as well as metrics that measure how well the foreground and background content is preserved during the edit.
Results show that our method outperforms existing methods quantitatively while being general enough to perform various kinds of edits.
To sum up, our main contributions are:
\begin{packed_itemize}
    \item A unified image editing approach that formulates common 2D and 3D editing operations as geometric transformations of parts of the image.
    \item \coolname, a zero-shot optimization-based approach that incorporates geometric transforms directly within the attention layers of diffusion models enabling realistic edits while preserving object style.
    \item Qualitative results of 2D and 3D object edits enabled by our method without model fine-tuning (see Fig. \ref{fig:teaser}).
\end{packed_itemize}


\section{Related Work}
%
Image editing has been widely studied in computer vision and encapsulates a range of operations, such as object removal and addition~\cite{song2023objectstitch, avrahami2022blended}, style transfer~\cite{goodfellow2014generative,karras2019style,jing2019neural, hertz2023style}, and 2D and 3D transforms~\cite{kholgade20143d}.
One challenge with this problem is to keep the edit consistent within the \emph{global} context of the image.
Traditional methods such as Poisson image editing~\cite{poissonedit} use gradients of the context to blend edits with existing pixels, while inpainting methods uses boundary and context to fill in pixels~\cite{Yang_2017_CVPR}.
We discuss generative model-based and 3D-aware editing methods below.

\parahead{Text-Guided Image Editing}
%
%
There are several works using generative image models to edit images via changes to the text prompt.
The preservation of subject identity in different settings can be achieved by textual inversion along with additional losses to finetune the generative model~\cite{ruiz2022dreambooth}.
\textit{Null-text} inversion is an inversion approach where a null-text embedding is optimized to match an inverted noise trajectory for a given input image along with attention reweighting~\cite{mokady2022nulltext}.
Instead of an inversion process, text prompt edits can also be achieved by swap, or re-weighting of cross-attention maps derived from the visual and textual representation~\cite{hertz2022prompt}.
Edits with text prompts can also be achieved by using cross-attention from different prompts to manipulate self-attention maps~\cite{cao2023masactrl}.
Leveraging existing text-to-image models along with~\cite{brooks2022instructpix2pix} gives the ability to generate paired data for finetuning a generative model to achieve text-guided editing results.
These methods mostly produce images with style changes or enhancements, or object replacement.
\cite{epstein2024diffusion} leverage prompts and self guidance to perform 2D image edits of scaling and translation.
However, it is difficult to guide the diffusion model to perform a specific 3D geometric transform based on a prompt.
We extend the above approaches to build a method to handle geometric transforms without any additional training.

\parahead{Non-Text-Guided Image Editing}
%
Text-guided edits are mostly limited to appearance and style changes.
Non-text-guided edits on the other hand, can achieve a variety of edits.
Point-based editing approaches can perform local image edits.
\cite{shi2023dragdiffusion} propose a motion supervised latent optimization between the reference and target edit, to guide the denoising to obtain the edit while preserving the object identity.
Stroke-based editing can edit larger image regions, or even entire images~\cite{meng2022sdedit}, by projecting strokes onto the image manifold via diffusion.
For these methods, edits such as translations are however not possible.
ObjectStitch~\cite{song2023objectstitch} along with inpainting can achieve translation where the denoising diffusion is applied to a target asked region, and guided by the embedding of the object to stitch.
However, object style preservation is difficult in this setting.
Recent methods~\cite{mou2023dragondiffusion, mou2024diffeditor} try to preserve identity and allow for translations while requiring no training.
However, these are limited to 2D translations and scaling.
An editing approach which first `lifts' the object from a background is proposed in~\cite{pandey2023diffusion}.
The background is inpainted and a depth-to-image generative model is used, which performs the denoising conditioned on an input depth. 
However, this approach needs an additional text prompt while ours does not.
Additionally, we support various kinds of edits and not just 3D transforms.
\cite{motion_guidance2024} uses flow-guidance for image editing.
However, optical flow can be much harder to obtain compared to depth~\cite{yang2024depth}.
We present a method that performs 2D and 3D edits using precise geometric transformations while preserving identity and not requiring additional user input.

\parahead{3D-Aware Editing}
Some methods have addressed the 3D editing problem~\cite{kholgade20143d} by `lifting' objects into 3D and use 3D meshes and scene illumination to allow for proper blending of the edited object with the existing image context.
Other methods use NeRF~\cite{yuan2022nerf, yu2023edit, wang2022clip, dong2024vica, haque2023instruct} or works~\cite{liu2023zero, liu2023one} learn over large-scale datasets~\cite{deitke2023objaverse}, leverage geometry representations to perform edits but require multi-view images that are difficult to obtain.
Edits are also directly applied to generative models, \eg,~\cite{pan2023_DragGAN} propose a point-based edit along with motion supervision to guide the neighboring pixels.
The authors of~\cite{Niemeyer2020GIRAFFE} propose to represent foreground objects and background as neural feature fields, which can be edited and composited for a final output.
The method of~\cite{ling2023freedrag} addresses limitations of point-based editing in GANs, using template features rather than points for better tracking, and restricting search area around pixels to lines.

\parahead{Concurrent Works} 
\cbEdit{Concurrent works Instadrag~\cite{Shi2024InstaDragLF} \& MagicFixup~\cite{Alzayer2024MagicFS} perform drag-based edits by tedious training over large-scale video data, but they don't bake in geometry of the object within the architecture. 
However, they may allow better inpainting and novel view synthesis of objects (discussion in supplement). 
 }
\vspace{-0.3cm}
\section{Background}
%
%
\parahead{Denoising Diffusion}
We first briefly describe the concept of Denoising Diffusion Probabilistic Models (DDPM) used successfully by diffusion models for image generation~\cite{Rombach_2022_CVPR}. Images can be considered as samples drawn from a data distribution $q(x)$. Denoising diffusion aims to learn a parameterized model $p_{\theta}(x)$ that approximates $q(x)$ and from which new samples can be drawn. The (forward) diffusion process iteratively adds noise to an input image $x_0$, with $t=0$, according to either a fixed or learned schedule, represented by $\alpha_t$ with $t \in[1,T]$.
At each timestep, the latent encoding is performed according to a Gaussian distribution centered around the output of the previous timestep: $q(x_t|x_{t-1}) = \mathcal{N}({x}_t;\sqrt{\alpha}{x}_{t-1}, (1 - \alpha_t)\mathbf{I})$.
The parameters vary over time such that $p_\theta(x_T) := \mathcal{N}(\mathbf{0},\mathbf{I})$. Using the reparameterization trick, the noised version of input $x_0$ can directly be expressed as: $x_t = \sqrt{\bar{\alpha}_t}x_0 + \sqrt{1 - \bar{\alpha}_t}\epsilon_0$.

The reverse process, where noise is gradually removed at each step, can be expressed as the joint distribution $p_\theta(x_{0:T}) = p_\theta(x_T) \prod^T_{t=1} p^{(t)}_\theta(x_{t-1}|x_t)$. Under the assumption of trainable means and fixed variances, a neural network $\hat{\mathbf{\epsilon}}_\theta(x_t, t)$ can be trained with the objective of minimizing a variational lower bound to estimate the source noise $\epsilon_0 \sim \mathcal{N}(\epsilon; \mathbf{0}, \mathbf{I})$ that determines $x_t$ from $x_0$:
$L_\gamma(\epsilon_\theta) := \sum_{t=1}^T \gamma_t \mathbb{E}_{x_0 \sim q(x_0), \epsilon_t \sim \mathcal{N}(\mathbf{0}, \mathbf{I})}\left[ \parallel \epsilon_0 - \hat{\epsilon}_\theta(x_t, t) \parallel^{2}_{2} \right]$.
For more details see~\cite{luo2022understanding}.

\parahead{Conditioning and Efficiency}
This formulation can be extended to the conditional case, \ie,~$p_\theta(x_{0:T}|y)$.
The condition $y$ could be images, (encoded) text, or something else.
The computational bottleneck is the number of denoising timesteps $T$, however a non-Markovian variant Denoising Diffusion Implicit Models (DDIM) was introduced to reduce the number of timesteps~\cite{song2021denoising}.
To further reduce the computational burden, the diffusion process for images can be performed in a lower dimensional latent (feature) space, as proposed by~\cite{Rombach_2022_CVPR}.
A perceptually optimized pretrained decoder takes the latent $x_1$, and reconstructs the image $x_0$.
In our work, we use a latent diffusion model together with Classifier-Free Guidance (CFG)~\cite{ho2022classifierfree} for text conditioning.




\parahead{Attention}
Attention was introduced as an alternative to recurrent networks and large receptive fields in convolution-based neural networks, for capturing local and global context~\cite{attention_all_you_need2017, DosovitskiyB0WZ21}.
The scaled dot-product self-attention mechanism adopted in transformers has found widespread application in computer vision applications.
The input is a tuple [Query ($Q$), Key ($K$), Value ($V$)], each with learnable parameters via linear layers.
An attention layer constructs an attention map $\text{AM}(Q, K) := \text{Softmax}\left(\frac{QK^{T}}{\sqrt{d}}\right) \label{eq:sm_attn}$ and then computes attention as: $\text{Attention}(Q, K, V) := \text{AM}(Q, K)V \label{eq:sdp-attn}$. Here, $d$ is the dimension of the embedding.

In addition to self-attention, the query can be derived from another input source, \eg,~another modality, and using the key and values from the first input, the cross-attention between the two inputs can be computed via ~\Cref{eq:sm_attn} and~\Cref{eq:sdp-attn}.
An example of cross-attention is the activation of a word in a sentence with pixels in an image.

The correlation between semantics and pixels for image-text cross-attention can be modified in the denoising diffusion generative image setting to adjust the appearance of a given generated image~\cite{hertz2022prompt}. In addition, deriving masks from cross-attention to guide self-attention~\cite{cao2023masactrl} provides the ability to change the appearance of objects while maintaining object identity. 

\begin{figure*}[tb]
  \centering
  \vspace{-0.1in}
  \includegraphics[trim={1cm 0.5cm 1cm 0.5cm},clip, width=\textwidth]{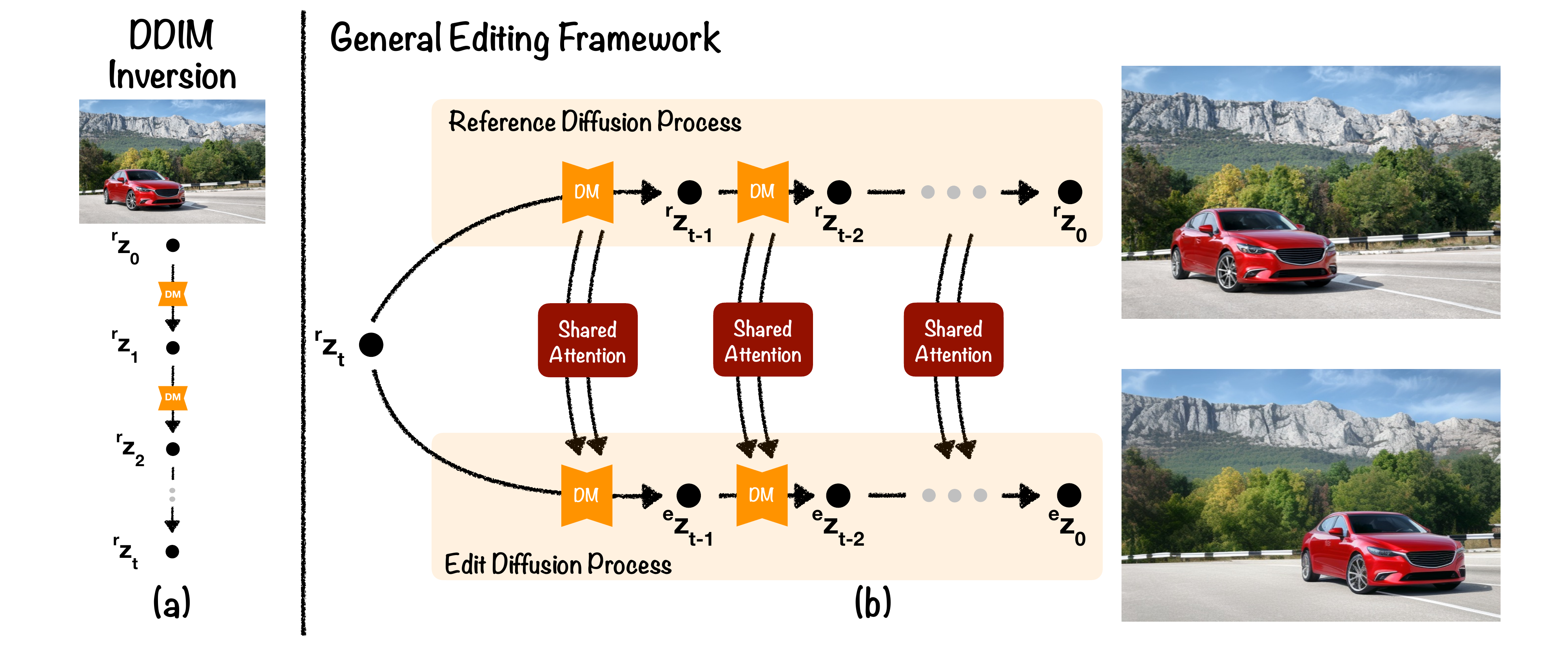}
  \vspace{-0.35in}
  \caption{
  General image editing framework using diffusion models.
  \textbf{(a) DDIM Inversion:} The process of obtaining noise trajectory $\{{}^{r}z_{0}, {}^{r}z_{1}, ......, {}^{r}z_{t}\}$ for the reference image~\cite{song2021denoising}.
  \textbf{(b) General Editing Framework:} The Reference Diffusion Process guides the Edit Diffusion Process to achieve the desired edit.
  In \textbf{\coolname}, we perform \textit{geometric} 2D and 3D edits by transforming the shared attention layers leading to plausible edits that preserves object style, inpainting disoccluded background, and adding details (\eg,~ the car's shadow).
  }
  \vspace{-0.25in}
  \label{fig:background}
\end{figure*}

\parahead{General Editing Framework}
Prior works leverage the learned capabilities of diffusion models to perform edits to a given image, rather than a generated one. A general framework (see \Cref{fig:background}) that is followed in all these works is to first perform an inversion \cite{mokady2022nulltext, song2021denoising} on the image.
This inversion provides us with a noise latent that sets a good starting point to regenerate the input image as well as to edit it.
Starting from the inverted noise latent, two parallel diffusion processes generate the input image as well as the edited image.
The first \textbf{reference diffusion process} generates the original input and, in our work, helps preserve un-edited regions of the image.
An \textbf{edit diffusion process} runs in parallel that utilizes the attention blocks from the reference process to perform the desired edit. This \textbf{shared attention} is a key insight for our proposed work. The editing framework is sketched in \Cref{fig:background} (b).

\begin{figure*}[tb]
  \centering
  \vspace{-0.15in}
  \includegraphics[width=0.98\textwidth, trim={2cm 0 0 0}, clip]{images/attention_sharing/Attention_sharing_lowest_fixed_camera.pdf}
  \vspace{-0.12in}
  \caption{\textbf{(a) \coolname} attention sharing mechanism that leverages the geometric transformation $\mathcal{F}(\cdot)$ transform the reference attention ${{}^{\mathcal{G}}Y_{ref}}$ to  guide the edit attention layer. \textbf{(b) Optimization Loss Functions} that penalize the latents and text-embeddings to perform the desired geometric edit. The orange mask highlights the region to be inpainted in the optimization.
  }
  \vspace{-0.25in}
  \label{fig:pipeline}
\end{figure*}
\section{\coolname{}}
The goal of \coolname is to enable editing of segmented foreground objects
in either natural or generated images.
We take the view that common editing operations like 2D translation, 3D object rotation or object removal can be expressed as geometric transformations of parts of the image.
Naively applying this transform to segmented foreground objects typically produces poor results w.r.t. image context and does not fill in the disoccluded background.
We propose to use diffusion models to realistically edit the image and preserve object style.

\parahead{Supported Operations}
In this paper, we focus on \emph{geometric edits} to an image $\mathcal{I}$ specified by users through sliders that control transformations of foreground objects.
In particular, we unify three kinds of edit operations that previously required separate bespoke methods: 
(1)~\textbf{2D object edit} operations deal with realistically translating or scaling segmented objects within the image including inpainting the background where the object was originally located.
(2)~\textbf{3D object edit} operations deal with realistically transforming objects based on user-specified 3D rotation, translation or scaling and inpainting any disoccluded background as a result of the edit.
Finally, (3)~\textbf{object removal} refers to the operation of removing the segmented object completely and inpainting the disoccluded background.

In contrast with previous approaches, we formulate edits as an optimization problem based on the shared attention and leverage a pre-trained text-to-image Stable Diffusion model~\cite{Rombach_2022_CVPR} to perform the edit.
Notably, our method requires no training and can use any diffusion model with attention.
Given an image $\mathcal{I}$, an object mask $M_{\text{obj}}$, a user-specified 2D or 3D transformation $T$, our goal is to edit the object in the image and inpaint any disoccluded regions introduced by the edit.
To compute $T$ for 3D edits, we use a depth map $D$ obtained from DepthAnything~\cite{yang2024depth} or simply by setting a constant depth of 0.5~m.
This enables us to edit in-the-wild natural images without additional user input.


\subsection{Edits via Shared Attention}
%
Each edit operation begins by performing a DDIM inversion~\cite{song2021denoising} on the given image (\Cref{fig:background} (a)).
Inverting the image provides us with the latent noise trajectory that will guide the edit diffusion process.
We then perform the reverse diffusion process along with the geometry-aware attention sharing mechanism as sketched in~\Cref{fig:background} (b).
This attention sharing mechanism along with optimizing for the image latents as well as text embeddings is the key to achieve the desired geometric edit.~\Cref{fig:pipeline} (a) depicts the process for the shared attention blocks from~\Cref{fig:background} (b).

\parahead{Image Inversion}
For inversion, we use direct inversion~\cite{ju2023direct} on the image $\mathcal{I}$ with the null prompt "". 
\cbEdit{Direct inversion initializes the reference trajectory with the noising trajectory for fast and accurate reconstruction of the reference image without the need for optimizing embeddings (null-text \cite{mokady2022nulltext}) \& model weights (LoRA \cite{shi2023dragdiffusion})}.
This inversion provides us with latents $\{{}^{\text{r}}z_t, {}^{\text{r}}z_{t-1}... {}^{\text{r}}z_0\}$ that preserves the style of the image and guides the edit.

\parahead{2D Edits}
\coolname can perform 2D edits without requiring a depth map. 
Through a user interface, we can obtain a transformation $T$ corresponding to a desired 2D translation or scaling. 
We define a 2D signal $S: [0, 1]^{2} \rightarrow \mathbb{R}^{C}$ that stores a per-pixel feature in the image. 
The signal $S$ can represent the RGB values or even the features of a deep network defined at each coordinate. 
Given a per-pixel edit $\mathcal{F}$ defined on $S$, our shared attention mechanism uses $\mathcal{F}$ to transform this signal for the desired edit. 
In our case, this signal is the Query embedding of the attention layer.

%

\parahead{3D Edits}
%
2D edits are limited as they do not leverage the geometry of objects.
We can extend 2D edits to 3D by additionally incorporating depth information $D$ monocular depth estimators~\cite{yang2024depth, bhat2023zoedepth} or simply a constant billboard depth map.
The user specifies a 3D rigid transformation $T$ which can then be used to compute the per-pixel edit $\mathcal{F}$ as 
\begin{equation*}
    \mathcal{F}(S)[u] := S[PTD[u]P^{-1}u].
\end{equation*}
Here, $P$ is the camera intrinsic matrix that is used to project points in the image and $u$ is the coordinate location of the signal. 
This edit field $\mathcal{F}$ captures the 3D shape of the visible region of the object and provides an estimate of the desired location of the object.
Note that if the per-pixel edit field is known, \eg,~from optical flow, we do not need a depth map for guidance.
However, optical flow is much more challenging to obtain for a single image compared to depth maps.


\parahead{Object Removal}
Object removal introduces disocclusions to the background where the object was originally located.
We propose an additional loss (see \Cref{sec:opt}) for the optimization of the diffusion latents to handle such disocclusions.
Disocclusions can also occur for 2D and 3D edits, and we consider such edits to be composites of removal and placement operations.
Our proposed formulation for latent optimization thus extends to those edits as well.  


\parahead{Shared Attention}
A key insight of our work is that we can transform objects by merely applying the edit $\mathcal{F}$ to the query embeddings of the reference attention (\Cref{fig:pipeline} (a)). Let ${}^{\text{r}}Q, {}^{\text{r}}K, {}^{\text{r}}V$ be the queries, keys, and values within the diffusion model of the reference denoising process and ${}^{\text{e}}Q, {}^{\text{e}}K, {}^{\text{e}}V$ be the queries, keys, and values of the corresponding attention block in the edit denoising process. 
The reference attention guidance ${}^{\mathcal{G}}Y_{\text{ref}}$ and edit attention guidance ${}^{\mathcal{G}}Y_{\text{edit}}$ are then given by
%
\begin{align}
    & {}^{\mathcal{G}}Y_{\text{ref}} := \text{Attention}(\mathcal{F}({}^{\text{r}}Q), {}^{\text{r}}K, {}^{\text{r}}V) \\
    & {}^{\mathcal{G}}Y_{\text{edit}} := 
     \begin{cases}
        \text{Attention}({}^{\text{e}}Q, {}^{\text{r}}K, {}^{\text{r}}V),  \text{if } \text{SelfAttention} \\
        \text{Attention}({}^{\text{e}}Q, {}^{\text{e}}K, {}^{\text{r}}V),  \text{otherwise} 
    \end{cases}
    \label{eq:cross-attn-edit}
\end{align}

%
\cbEdit{Applying transformation $\mathcal{F}$ only to the reference query embeddings ${}^rQ$ followed by dot product with the reference key embeddings ${}^rK$ in Eq.~\ref{eq:cross-attn-edit} provides us with correspondences between them. 
This edited attention map attends to the reference value embeddings and ensures that the transform only changes the geometry and preserves the appearance.
We use the edit key embeddings ${}^eK$ in cross attention map to enable the flow of gradients to the null-initialized text embeddings for the optimizing the edit trajectory.}
To place the object at the desired location, the edit and reference attention guidance should approximately be the same (${}^{\mathcal{G}}Y_{\text{ref}} \approx {}^{\mathcal{G}}Y_{\text{edit}}$) for the foreground. 
Note that they need not be exactly the same in the case of an ill-defined edit $\mathcal{F}$. We then transform the output ${}^{\mathcal{O}}Y_{\text{edit}}$ of the edit attention layer
\begin{align}
    & {}^{\mathcal{O}}Y_{\text{edit}} := \mathcal{F}(M_{obj}) \cdot {}^{\mathcal{G}}Y_{\text{ref}} + (1 - \mathcal{F}(M_{obj})) \cdot {}^{\mathcal{G}}Y_{\text{edit}},
    \label{eq:edit-guidance}
\end{align}
where $\mathcal{F}(M_{obj})$ refers to the foreground mask after applying the transformation $\mathcal{F}$. In other words, Eq.~\ref{eq:edit-guidance} aim to preserve identity for the object in the edit at its target location, while simultaneously preserve identity and consistency for the remaining pixels (or background). See supplement for algorithm and details.



\subsection{Optimization}
\label{sec:opt}
\coolname is a zero-shot optimization-based method that operates \textbf{without the need for any additional training}. 
We achieve this via optimization of the latents and null-initialized text embeddings for edit guidance. 
The shared attention guidance provides us with a proxy of where the foreground object must be placed after the edit. 
\cbEdit{However, it does not guide the inpainting of the disocclusions introduced by moving the object causing duplications.
We formulate an optimization procedure to fill the disocclusions and penalize the deviation of the edit attention guidance from the reference attention guidance.} 
The loss functions used in the optimization (shown in Fig. \ref{fig:pipeline} (b)) are explained in detail next.



\parahead{Background Preservation Loss}
Performing shared attention guidance along with optimization could result in the un-edited regions of the image to also be changed.  
We introduce a background preservation loss to prevent this.
Let the mask $M_{\text{ne}}$ represent the non-editable region of the image.
We define the background preservation loss as
\begin{align}
    \mathcal{L}_{bg} := \text{mean}(M_{\text{ne}} \cdot ||{}^{\mathcal{G}}Y_{\text{edit}} - Y_{\text{ref}}||_1).
\end{align}
Here,  {\small $Y_{\text{ref}} = \text{Attention}({}^{\text{r}}Q, {}^{\text{r}}K, {}^{\text{r}}V)$} is the attention block output for the reference de-noising process. The reference attention preserves the style of the image and constrains the optimization towards preserving the background.

\parahead{Object Placement Loss}
Occasionally, the optimization changes the foreground region of the image.
This causes loss of detail in the foreground. 
To prevent this, we penalize the deviation between the edit guidance and the reference guidance within the transformed foreground mask by
\begin{align}
    \mathcal{L}_{obj} := \text{mean}(\mathcal{F}(M_{\text{obj}}) \cdot ||{}^{\mathcal{G}}Y_{\text{edit}} - {}^{G}Y_{\text{ref}}||_1).
\end{align}
%
Note, we don't use this loss for object removal. 
%



%

\parahead{Inpainting Loss}
To inpaint the disoccluded regions of the image, we maximize the difference between the edit guidance attention map {\small ${}^{\mathcal{G}}A_{\text{edit}} := \text{AM}({}^{\text{e}}Q, {}^{\text{r}}K)$} and the reference guidance attention map $A_{\text{ref}}:= \text{AM}({}^{\text{r}}Q, {}^{\text{r}}K)$.  
Let $\rho_{\text{obj}\rightarrow \text{bg}}$ represent the maximum normalized correlation score for each row in the foreground mask of the attention map ${}^{\mathcal{G}}A_{\text{edit}}$ to each row in the background mask of the reference attention map $A_{\text{ref}}$.
We can similarly compute $\rho_{\text{obj}\rightarrow \text{obj}}$ that provides us with the maximum foreground to foreground normalized correlation (see \Cref{fig:pipeline} (b)). 
Our goal is to reduce $\rho_{\text{obj}\rightarrow \text{obj}}$ and increase $\rho_{\text{obj}\rightarrow \text{bg}}$.
We want to inject the disoccluded region with features from the background and ensure that the diffusion process doesn't in-paint the same features. We penalize for this using
\small
\begin{footnotesize}
\begin{align}
    \mathcal{L}_{remove} := \text{mean} \left( e^{-d_{\text{obj}\rightarrow \text{bg}}} (ln(\rho_{\text{obj}\rightarrow \text{obj}}) - ln(\rho_{\text{obj}\rightarrow \text{bg}}))\right).
\end{align}
\end{footnotesize}
\normalsize
Here, $d_{\text{obj}\rightarrow \text{bg}}$ is the coordinate distance between the locations of the attention map.
The loss weighted by coordinate distance ensures that the foreground region inpaints the region using features within its vicinity.
The negative log forces the object to background correlation $\rho_{\text{obj}\rightarrow \text{bg}}$ to increase and also reduces object-object correlation forcing the inpainted region to not be filled by the same object.



\parahead{Smoothness Constraint}
We additionally penalize the edit attention guidance ${}^{\mathcal{G}}Y_{\text{edit}}$ for smoothness by penalizing the absolute value of its gradients using $\mathcal{L}_{s}$.

\parahead{Geometry Editing Optimization}
\cbEdit{
We edit images by penalizing the null-initialized text embeddings and images latents during generation using the final loss $\mathcal{L} := w_{bg}\mathcal{L}_{bg} + w_{obj}\mathcal{L}_{obj} + w_{r}\mathcal{L}_{remove} + w_{s}\mathcal{L}_{s}$.
}

In our experiments, we found that the inpainting loss $\mathcal{L}_{remove}$ is hard to optimize and changes every image differently. 
To combat this, we devise an adaptive optimization scheme that increases the weight $w_r$ of the removal loss if the loss is more than -1.8 and reduce the loss weight if the removal loss is lower than -6. 
\cbEdit{All our experiments are performed on an Nvidia RTX3090 with a run time of 30 seconds (for removal) up to 45 seconds (for 2D \& 3D edits) on image resolution of 512.} 
We also penalize depth smearing artefacts of the foreground using amodal loss $\mathcal{L}_{amodal}$.
See supplement for algorithm \& more details.
\section{Results \& Experiments}
\label{sec:expt}
In this section, we present visual examples of our editing results and quantitative results of visual metrics of editing quality and a perceptual study.

\parahead{Dataset}
To measure the efficacy of our method we collected a dataset of real images from Adobe Stock images \cite{adobe-stock} to ensure we exclude generative AI images.
We collect 70 images corresponding to the prompts \textit{dog, car, cat, bear, mug, lamp, boat, plane, living-room, peaceful scenery}. 
We also test on real in-the-wild images from~\cite{geiger2013vision} and generated images from \cite{pandey2023diffusion}.
For many images in our dataset, we show multiple 2D and 3D edits demonstrating the general editing capabilities of \coolname.

\parahead{Baselines}
Since there is no extant method that performs all types of edits that we support, we compare each edit type to a different baseline. 
For the object removal, we compare with a state-of-the-art off-the-shelf \textbf{LaMa} image in-painting model~\cite{lama-inpainting}, dilating object masks to make LaMa work better.
For the 3D edit operations, we benchmark ourselves against \textbf{Zero123-XL}~\cite{liu2023zero}, \textbf{FreeDrag}~\cite{ling2023freedrag}, \textbf{DragonDiffusion}~\cite{mou2023dragondiffusion, mou2024diffeditor}, \textbf{Diffusion Self Guidance} ~\cite{epstein2024diffusion} and \textbf{Diffusion Handles}~\cite{pandey2023diffusion}. Please see supplement for more details about each baseline.
Moreover, our tests with \textbf{Object 3DIT}~\cite{object-edit} on real images produced poor results so we exclude it.
For 2D edits, we compare with \textbf{Dragon Diffusion} \cite{mou2023dragondiffusion, mou2024diffeditor}.
Since above methods require prompts while our method does not, we manually added text descriptions to our data.
We benchmark our method against baselines using community accepted metrics of Mean Distance, Clip Similarity Score, and Warp Error. 
We additionally test our method for inpainting and editing with a perceptual study.

\subsection{Quantitative Evaluation}

\parahead{Metrics}
We detail the metrics here to evaluate our edits quantitatively against baselines for edit adherence and style preservation. 
We performed a total of 102 edits on our dataset: 36 2D edits and 66 3D edits.
Metrics such as FID and Image Fidelity (IF)~\cite{shi2023dragdiffusion,mou2023dragondiffusion} are not suitable for evaluating geometric edits because there could be large visual difference (e.g., large translation) and they do not measure disocclusion inpainting quality.
\begin{figure}
\vspace{-0.05in}
    \centering
    \scalebox{0.75}{
\setlength{\tabcolsep}{6pt} 
\renewcommand{\arraystretch}{1} 
\begin{tabular}{l|rrrrr}
    \toprule
     &\textbf{MD} {\color{blue} $\downarrow$}  & \multicolumn{1}{c}{\begin{tabular}[c]{@{}c@{}}\textbf{Warp} \\ \textbf{Error}\end{tabular}} {\color{blue} $\downarrow$} & \begin{tabular}[c]{@{}c@{}}\textbf{Clip}\\ \textbf{Similarity}\end{tabular}$\uparrow$  & & \\
    \midrule
    \multicolumn{3}{l}{\textbf{3D Edits}} \\
    \midrule
    \textbf{\cbEdit{Diffusion Self Guidance} \cite{epstein2024diffusion}} & 92.067 & 0.243 & 0.809\\
    \textbf{Dragon Diffusion \cite{mou2023dragondiffusion, mou2024diffeditor}} & 66.108  & 0.226 & 0.953 \\
    \textbf{FreeDrag \cite{ling2023freedrag}} & 31.451 & 0.182 &  \textbf{0.977}\\
    \textbf{Zero123-XL + Lama \cite{liu2023one, lama-inpainting}} & 19.010&  0.157 & 0.961\\
    \textbf{Diffusion Handles \cite{pandey2023diffusion}} & 10.837 & 0.114 &  0.890\\
    \textbf{\coolname (Ours)} & \textbf{7.304}& \textbf{0.091} & 0.967\\
    \midrule
    \multicolumn{3}{l}{\textbf{2D Edits }} \\
    \midrule
    \textbf{\cbEdit{Diffusion Self Guidance} \cite{epstein2024diffusion}} & 155.149 & 0.297 & 0.806 \\
    \textbf{FreeDrag \cite{ling2023freedrag}} & 64.716 & 0.259 & 0.962 \\
    \textbf{Zero123-XL + Lama \cite{liu2023one, lama-inpainting}} & 20.000 & 0.135 & 0.929\\
    \textbf{Dragon Diffusion \cite{mou2023dragondiffusion, mou2024diffeditor}} & 38.070& 0.151 & 0.957 \\
    \textbf{\coolname (Ours)} & \textbf{5.579} & \textbf{0.098} & \textbf{0.963} \\
    \bottomrule
\end{tabular}

}
    \vspace{-0.3cm}
    \captionof{table}{Our method adheres to the desired edit having the least \textbf{Mean Distance} and \textbf{Warp Error} compared to Dragon Diffusion, FreeDrag, Diffusion Self Guidance, and Diffusion Handles.
    \label{tab:metrics}}
\vspace{-0.7cm}
\end{figure}



Therefore, we use three other metrics to better evaluate methods:
(1)~The \textbf{Mean Distance (MD)} metric computes interest points on the foreground of the image using SIFT~\cite{sift} and finds correspondences between the input and edited image using DiFT~\cite{dift}.
We then measure the distance between the correspondence estimated by DiFT and the edit specified by the user. This metric measures how well each approach transforms the foreground object.
(2)~the \textbf{Warp Error (WE)} metric forward warps the foreground region of the input image to the edited image and compute the absolute difference between their pixels for the transformed foreground. 
This metric measures how well each approach adheres to the edit.
(3)~the \textbf{CLIP Similarity (CS)} metric computes the CLIP image embedding~\cite{wang2022clip} of the input and edited image and measures the cosine similarity.
A good editing approach preserves the image context with high \textbf{CS} and adheres to the edit with low \textbf{WE} \& \textbf{MD}.  

\Cref{tab:metrics} shows quantitative comparison for 2D and 3D edits of our method with the baselines.
\coolname (\textbf{MD(2D):} 5.579 \& \textbf{MD(3D)}: 7.304) outperforms baselines FreeDrag, Diffusion Handles, Zero123-XL, Dragon Diffusion, and Diffusion Self Guidance in \textbf{MD} metrics and Warp Error for both 2D and 3D edits.
Dragon Diffusion does not perform well in these tasks since their method fails to inpaint disocclusions or preserve the foreground.
Zero123-XL baseline performs better but since it is not trained on real-world images, it does not preserve the foreground object resulting in incorrect DiFT correspondences.
\cbEdit{All methods seem to preserve the context of the scene with a clip score above 0.920 apart from Diffusion Handles and Diffusion Self Guidance that struggle to preserve the image style in 3D edits with clip score of 0.890 \& 0.809 respectively.} But Diffusion Handles better adheres to the edit as it uses the depth to project activation functions of the SD model. 
\cbEdit{Diffusion Self Guidance consistently underperforms as it most often does not move objects and does not preserve the object appearance.}
For 3D edits, FreeDrag has a marginally high clip similarity score of 0.977 compared to ours (\textbf{CS:} 0.967). However, at times \textbf{CS} is higher when the foreground is not removed appropriately.
FreeDrag struggles to optimize for large edits and occasionally produces improper object inpainting. 
See \Cref{fig:comparison_3D} and supplement for visual comparison and a detailed implementation, timing, and performance analysis for all baselines. 



\parahead{Perceptual Study}
\cbEdit{In addition to quantitative evaluations, we perform a perceptual study with 53 participants to compare our inpainting and editing results against prior works.}
This was setup as a forced choice questionnaire where participants had to select one of two options as containing the best edit result.
Of the two randomly presented options, one was ours and the other was a baseline.
Participants preferred our inpainting over LaMa~\cite{lama-inpainting} 94.06\% of the time.
They also preferred our geometric edits over Zero123-XL 86.48\% of the time for realism and 88.48\% of the time for edit adherance.
See supplement for more information.

\begin{figure}[ht]
    \vspace{-0.3cm}
    \centering
\scalebox{0.65}{
\begingroup
\setlength{\tabcolsep}{6pt} 
\renewcommand{\arraystretch}{1} 
\begin{tabular}{l|rrr}
    \toprule
     &\textbf{MD} {\color{blue} $\downarrow$}  & \multicolumn{1}{c}{\begin{tabular}[c]{@{}c@{}}\textbf{Warp}\\ \textbf{Error {\color{blue} $\downarrow$}}\end{tabular}}  & \begin{tabular}[c]{@{}c@{}}\textbf{Clip}\\ \textbf{Similarity}\end{tabular}$\uparrow$  \\
    \midrule
    \multicolumn{3}{l}{\textbf{Timesteps for Geometric Attention Sharing (Geometry Guidance)}} \\
    \midrule
    \textbf{t=30} & 8.363 & 0.0998 & 0.872\\
    \textbf{t=37} & 7.158 & 0.0950 & 0.932\\
    \textbf{\coolname (t = 45)} & \textbf{6.785}& \textbf{0.0934} & \textbf{0.966}\\
    \midrule
    \multicolumn{3}{l}{\textbf{Adaptive Optimization }} \\
    \midrule
    \textbf{w/o Adaptive Optimization} &  9.164 &  0.0944& \textbf{0.966} \\
    \textbf{\coolname (with Adaptive Optimization)} & \textbf{6.785}& \textbf{0.0934} & \textbf{0.966} \\
    \midrule
    \multicolumn{3}{l}{\textbf{Loss Functions }} \\
    \midrule
    \textbf{w/o Background Preservation Loss} & \textbf{6.736} & 0.0958 & 0.954 \\
    \textbf{w/o Removal Loss} & 57.600 & 0.0941&  0.965\\
    \textbf{w/o Object Placement Loss} & 7.397& 0.0986 &  0.963\\
    \textbf{GeoDiffuser} & 6.785& \textbf{0.0934} & \textbf{0.966} \\
    \bottomrule
\end{tabular}
\endgroup
}
    \vspace{-0.2cm}
    \captionof{table}{\cbEdit{\textbf{Metric Ablations}: Increasing the number of time steps for geometric attention sharing and adaptive optimization both improve the Mean Distance, Warp Error, and Clip Similarity score. 
    Removing removal loss introduces duplication of objects and removing background preservation changes the scene background.}
\label{tab:ablation_metrics}}
\vspace{-0.3cm}

\end{figure}

\begin{figure*}[t]
\centering
\vspace{-0.3in}
\includegraphics[width=0.98\textwidth]{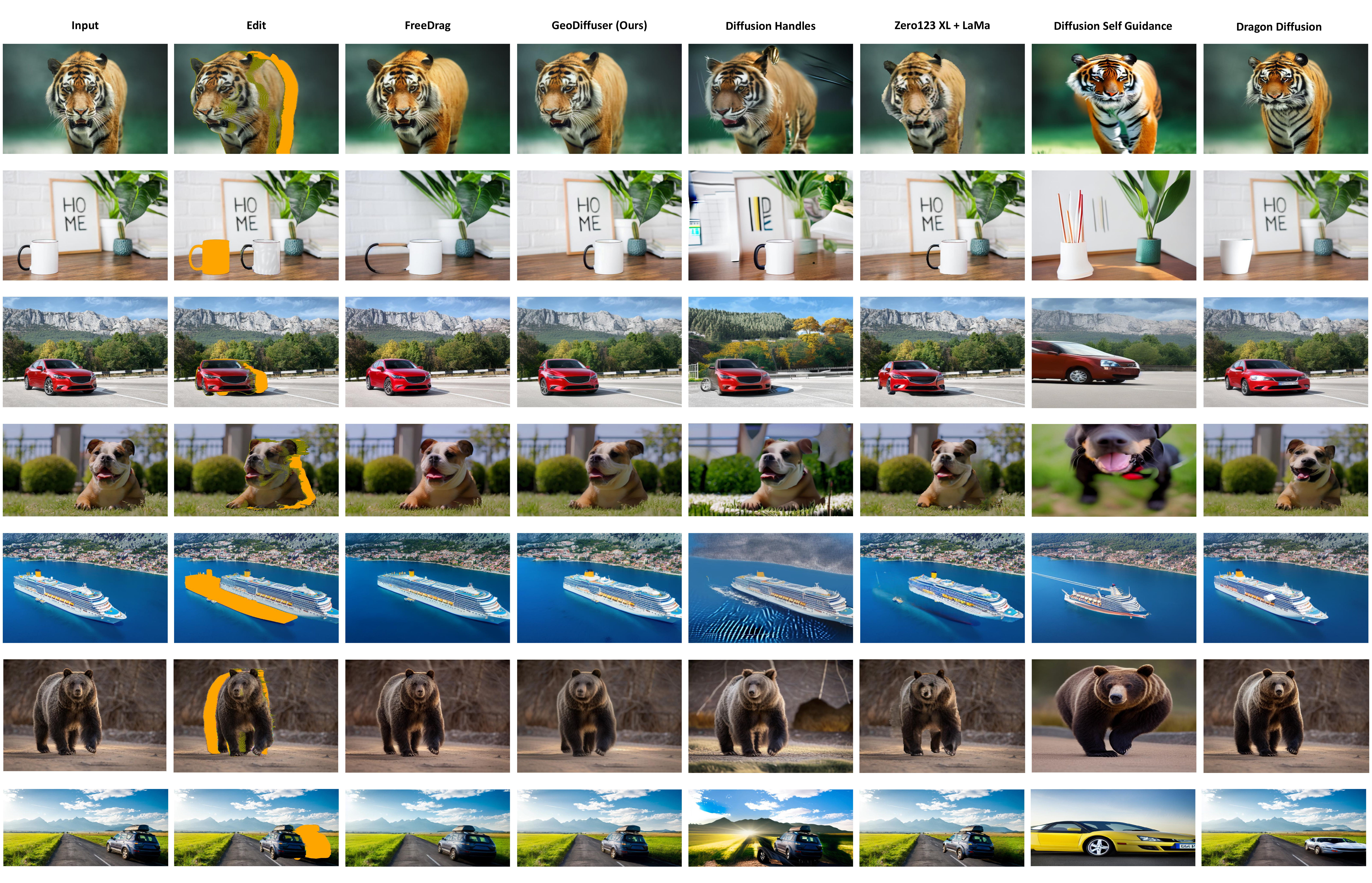}
\vspace{-0.2in}
\caption{\cbEdit{We perform the same edit using prior works and compare with out work. 
We show the intended 3D edit in column 2 where we highlight the region to be inpainted with orange and the region foreground inpainting region with green. 
Our work \coolname best adheres to the intended edit and ensures preservation of the scene without requiring prompts. 
Diffusion Handles requires an inpainting model and a depth trained diffusion model to perform the same edit with prompts but still fails to preserve the appearance of the scene. 
FreeDrag is slow and does not adhere well to the edit.
Dragon Diffusion and Diffusion Self Guidance do not preserve the appearance of the object and do not rotate objects accurately.
Please see supplement for a detailed analysis of all prior works.
}}
\vspace{-0.23in}
\label{fig:comparison_3D}
\end{figure*}
\parahead{Ablations} 
%
We present quantitative ablations of our design choices in \Cref{tab:ablation_metrics}. Increasing the number of time steps for geometric attention sharing provides geometric guidance for more accurate edits with lower \textbf{MD} and \textbf{WE} (\Cref{tab:ablation_metrics}, \Cref{fig:ablation_guidance}). 
Without adaptive optimization, we need image specific tuning for loss weights which is not scalable. 
\cbEdit{Removing placement loss reduces the foreground edit accuracy increasing the \textbf{MD} and \textbf{WE}. 
Background preservation loss improves scene preservation with improved global consistency and high \textbf{CS}.
Without removal loss there exist duplicates within the edited image that lead to incorrect correspondences while computing \textbf{MD} resulting in very high errors.
Please see supplement for more visual ablations.
}
\begin{figure}[H]
\vspace{-0.15in}
\centering
\scalebox{1.0}{
\centering
\includegraphics[width=0.48\textwidth]{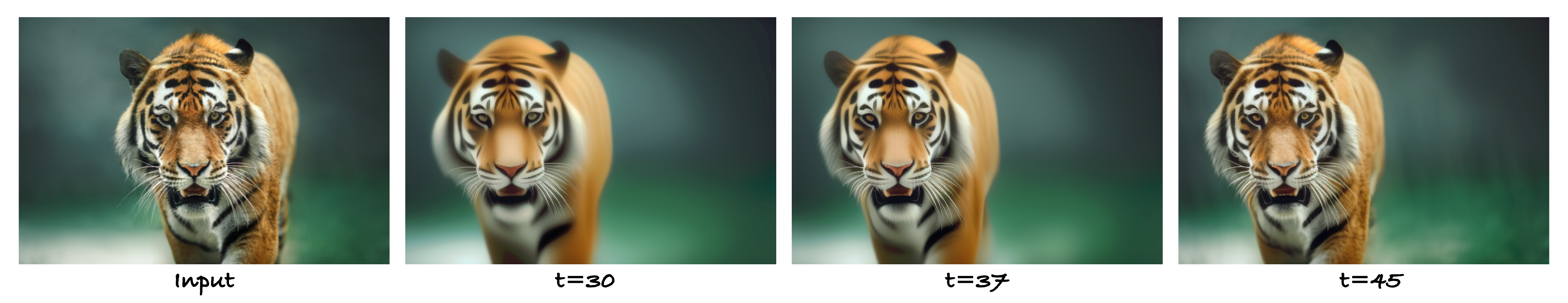}
}
\vspace{-0.3in}
\caption{\textbf{Geometry Guidance}:
Increasing steps $t$ for geometric attention sharing better preserves object style (translation edit).}
\label{fig:ablation_guidance}
\vspace{-0.2in}
\end{figure}


\parahead{Qualitative Results}
We show more qualitative comparisons of 3D edits performed by \coolname against baselines in \Cref{fig:comparison_3D} and supplement.
Note how \coolname not only removes / transforms objects but also their reflection and shadows.




\vspace{-0.3cm}
\section{Conclusion}
\coolname is a unified method to perform common 2D and 3D object edits on images.
Our approach is a zero-shot optimization-based method that uses diffusion models to achieve these edits.
The key insight is to formulate image editing as 
a geometric transformation and incorporate it directly within the shared attention layers in a diffusion model-based editing framework.
Results show that our single approach can handle a wide variety of image editing operations, producing better results compared to prior work.

\parahead{Limitations \& Future Work}
While we can handle background disocclusions, we cannot yet handle \cbEdit{foreground object disocclusions resulting from large 3D rotations that requires accurate novel view synthesis of in-the-wild objects which is a very difficult problem.}
Our method also occasionally generates artifacts due to downsampled attention masks and is limited by the capabilities of the base diffusion model (see supplement for details).
We plan to address these limitations in future work.

\parahead{Acknowledgements} 
Part of this work was done during Rahul’s internship at Amazon. 
This work was additionally supported by NSF grant CNS \#2038897, ONR grant N00014-22-1-259, and an AWS Cloud Credits award.

\appendix
\twocolumn[
\begin{center}{\bf {\Large 
Appendix 
}
}
\end{center}
]




\section{Qualitative Results}
We present more qualitative results towards the end of the document in \Cref{fig:gallery} and \Cref{fig:gallery_sup_1}. 
We also compare our method against prior works in \Cref{fig:comparison} for 2D edits.

\section{Implementation Details}

The shared attention along with the loss functions defined in the manuscript, enable performing geometry image edits as a reverse diffusion process by optimizing the latents and text embeddings. To make the optimization faster, we optimize every alternate step for the initial 32 diffusion steps. 
We set an initial learning rate of 1.5 and linearly decay it to 0. We share attention across all blocks within the UNet till step 45.
All our experiments are performed on an Nvidia RTX3090 with a run time of 30 seconds (for removal) up to 45 seconds (for 2D and 3D edits) on image resolution of $512$. 
Our timing is inclusive of the DDIM inversion, optimization with feature re-projection, and edit generation. 
We use \cite{ravi2020accelerating} for projecting, splatting, and rendering in our attention sharing mechanism. 
Occasionally, the histogram of the edited image does not match the input image and we match color histograms between the two.
We detail attention sharing mechanism in \Cref{algo:attention_sharing} and editing with \coolname in \Cref{algo:editing}.

\algrenewcommand\alglinenumber[1]{\footnotesize #1:}
\begin{algorithm*}[ht]
\footnotesize
\caption{Geometric Attention Sharing \label{algo:attention_sharing}}
\begin{algorithmic}[1]
\Require ${}^{\text{e}}Q (\text{\footnotesize edit query}), {}^{\text{e}}K (\text{\footnotesize edit key}), {}^{\text{r}}Q (\text{\footnotesize ref. query}), {}^{\text{r}}K (\text{\footnotesize ref. key}), {}^{\text{r}}V (\text{\footnotesize ref. value}), \mathcal{F} (\text{\footnotesize transformation}), M_{obj} (\text{\footnotesize object mask})$
\Ensure ${}^{\mathcal{O}}Y_{\text{edit}} := \text{AttentionSharing}({}^{\text{e}}Q, {}^{\text{e}}K, {}^{\text{r}}Q, {}^{\text{r}}K, {}^{\text{r}}V, \mathcal{F}, M_{obj})$
\State ${}^{\mathcal{G}}Y_{\text{ref}} := \text{Attention}(\mathcal{F}({}^{\text{r}}Q), {}^{\text{r}}K, {}^{\text{r}}V)$ \Comment{Reference Guidance and Applying Transform $\mathcal{F}$}
\If{\text{SelfAttention}} \Comment{If Self-attention block}
    \State ${}^{\mathcal{G}}Y_{\text{edit}} := \text{Attention}({}^{\text{e}}Q, {}^{\text{r}}K, {}^{\text{r}}V)$ \Comment{Use reference key}
\Else
    \State ${}^{\mathcal{G}}Y_{\text{edit}} := \text{Attention}({}^{\text{e}}Q, {}^{\text{e}}K, {}^{\text{r}}V)$ \Comment{Use edit key}
\EndIf
\If{\text{DiffusionCorrection}} \Comment{If Diffusion Correction (see \Cref{sec:supp_transform_correction})}
    \State ${}^{\mathcal{O}}Y_{\text{edit}} :={}^{\mathcal{G}}Y_{\text{edit}}$ \Comment{${}^{\mathcal{G}}Y_{\text{edit}}$ automatically finds correspondences between ${}^\text{e}Q$ and ${}^{\text{r}}K$ to correct the transformation enabling plausible edits.}
\Else
    \State ${}^{\mathcal{O}}Y_{\text{edit}} := \mathcal{F}(M_{obj}) \cdot {}^{\mathcal{G}}Y_{\text{ref}} + (1 - \mathcal{F}(M_{obj})) \cdot {}^{\mathcal{G}}Y_{\text{edit}}$
\EndIf
\State \Return ${}^{\mathcal{O}}Y_{\text{edit}}$
\end{algorithmic}
\end{algorithm*}

\algrenewcommand\alglinenumber[1]{\footnotesize #1:}
\begin{algorithm*}[ht]
\footnotesize
\caption{Geometric Editing with GeoDiffuser \label{algo:editing}}
\begin{algorithmic}[1]
\Require ${}^{r}z_{0} (\text{\footnotesize reference latent}), \mathcal{F} (\text{\footnotesize transformation}), M_{obj} (\text{\footnotesize object mask}), \Phi (\text{\footnotesize null-prompt or optional text})$
\Ensure ${}^{e}z := \text{GeometricEdit}({}^{r}z_{0}, \mathcal{F}, M_{obj}, \Phi)$
\State \{${}^{r}z_{T}, {}^{r}z_{T-1}..., {}^{r}z_{1} $\} $\leftarrow$ DDIMInversion(${}^{r}z_{0}$, $\Phi$) \Comment{Reference Inversion}
\State ${}^{e}z := {}^{r}z_{\text{T}}; {}^{r}z := {}^{r}z_{\text{T}}$ \Comment{Initialize edit latent with reference latent}
\For{t = T $\rightarrow$ 1} 
    \If{(t $\leq 30$) AND (t $\% 2 == 0$)} \Comment{Optimize}
        \State \_, \_, $\mathcal{L}_{dict}$  := DiffusionStep([${}^{r}z, {}^{e}z$], $\Phi, \mathcal{F}, M_{obj}, t$) \Comment{Diffusion Step with Attention Sharing and Loss Dictionary Computation}
        \State $\mathcal{L}$ := AdaptiveLoss($\mathcal{L}_{dict}$) \Comment{Weigh losses adaptively and sum}
        \State ${}^{e}z := {}^{e}z - \nabla_{{}^{e}z} \mathcal{L}$; $\Phi := \Phi -  \nabla_{\Phi} \mathcal{L}$ \Comment{Optimization Update by backpropagating through the diffusion model}
    \EndIf
    \State \_, ${}^{e}z$, \_ := DiffusionStep([${}^{r}z, {}^{e}z$], $\Phi, \mathcal{F}, M_{obj}, t$) 
    \State ${}^{r}z := {}^{r}z_{t-1}$ \Comment{Update reference latent with inversion trajectory for Direct Inversion \cite{ju2023direct}}
\EndFor
\State \Return ${}^{e}z$
\end{algorithmic}
\end{algorithm*}

\section{Evaluation and Baselines}
We detail the procedure to perform geometric edits using all our baselines. 
We also perform a timing and performance analysis of each baseline.

\subsection{FreeDrag \cite{ling2023freedrag}} 
\parahead{Implementation} FreeDrag  extends DragDiffusion \cite{shi2023dragdiffusion} to perform drag edits with better point tracking. 
We use the diffusion version in the official FreeDrag implementation which works better on real-world images for our evaluation.
For each edit, we first uniformly sample 40 points within the object mask and use the per pixel transform $\mathcal{F}$ to get the target points of the drag to edit images using FreeDrag. 
This ensures that the same geometric transform is used for editing for a fair comparison.
Sampling more points increased the edit time and did not improve the results. 
Each FreeDrag edit performs 200 LoRA steps with text prompt followed by 1000 drag optimization steps.  
We had to increase the optimization steps from 300 to 1000 in their implementation as FreeDrag did not converge correctly for large edits tested in our work with 300 steps.

\parahead{Timing Analysis} The 200 LoRa optimization steps runs for 116.26 seconds and the optimization using 1000 steps runs for 165.24 seconds.

\parahead{Performance Analysis} We notice that FreeDrag optimization averages nearby drag vectors and does not adhere to the edit. 
Additionally, it often stretches objects as it does not have removal capabilities baked into the optimization and does not track points appropriately for large edits while our method produces plausible results while being significantly faster (see \Cref{fig:comparison} and Fig. 4 \manu).

\subsection{Diffusion Handles \cite{pandey2023diffusion}} 
\parahead{Implementation} We use the official implementation from the authors of Diffusion Handles. 
Each edit utilizes the depth map and camera transformation to perform the geometric edit. 
Diffusion handles first performs a null-text inversion using the depth to image stable diffusion model and then inpaints the foreground region of the object using LaMa \cite{lama-inpainting}. 
The inpainted image is then used to estimate the background depth of the scene. 
The background depth is blended with the transformed foreground object.
This transformed depth map along with transformed activations of the depth to image SD model is then used to generate the edited image as detailed in \cite{pandey2023diffusion}.
Additionally, we had to change the camera FOV to 49.92$^{\circ}$ to ensure that the same transformation is applied during the edit. 

\parahead{Timing Analysis} Each edit requires 60 seconds of Null-text optimization followed by 35 seconds of edit. 

\parahead{Performance Analysis} We notice that \cite{pandey2023diffusion} fails to preserve the image content and style, but adheres to the foreground transformation well.
However, the image style is not preserved when the depth maps are not predicted using \cite{midas_Ranftl2019TowardsRM} because they are not in the training distribution of Depth to Image Stable Diffusion model.
This leads to low Clip Similarity (CS) and degradation in content preservation as shown in the qualitative comparisons if our manuscript.
However, we do not have this limitation and can leverage depth maps from any monocular depth estimator.
Another limitation for Diffusion Handles is the reliance on multiple depth predictions (for foreground as well as background) and then merging the foreground depth with the background depth. 
The image generated using this merged depth map produces improper object removal and at times replaces the object with another instance of the same type. 
2D edits with \cite{pandey2023diffusion} were not good as a constant depth for foreground object was not producing good results even after null-text optimization.

\subsection{Dragon Diffusion \cite{mou2023dragondiffusion, mou2024diffeditor}} 
\parahead{Implementation} We use the 2D movement feature of the official Dragon Diffusion implementation for 2D edits and 40 drag points for 3D edits. 
We use the camera projection, mask, and depth maps to get the target point locations similar to the FreeDrag implementation.
We also tried using 100 drag points to perform 3D edits, but this made results worse as the edit moved objects partially, introduced holes, and did not preserve its appearance.
For 2D edits, its movement feature utilizes an object mask, a source point and a target drag location.
We use the IP adapter ~\cite{mou2024diffeditor} that is trained for editing as well for this benchmark, but it did not edit real images very well.
We had to increase the weights for $\epsilon_{opt}$ and $\epsilon_{content}$ losses for better object removal and content preservation to perform real-world edits.

\parahead{Timing Analysis} Dragon Diffusion performs inversion in 4 seconds and uses an optimized implementation that edits images in 20 seconds.
This method is quick as it doesn't deal with 3D geometry projection and uses the memory bank to speed up the generation process.
We can leverage the memory bank to speed up our model as a future work.

\parahead{Performance Analysis} Dragon Diffusion does not perform well to inpaint disocclusions or preserve the foreground.
It has a marginally high clip similarity score as it does not completely remove the object from the source introducing duplicates.

\subsection{Zero123-XL + LaMa~\cite{liu2023zero, lama-inpainting}}
\parahead{Implementation} For this baseline, we first use~\cite{lama-inpainting} to in-paint the region of the removed foreground object.
We then Zero123-XL to predict the novel view of the transformed object and composite it to the in-painted background image using Laplacian pyramid blending.

\parahead{Timing Analysis} Zero123-XL + LaMa takes about 5 seconds to run for each edit

\parahead{Performance Analysis} Zero123-XL moves the object and LaMa removes the object, but it fails to preserve the foreground accurately as it is not in the model's training distribution.
It is also difficult to control the per-pixel transform accurately with Zero123-XL as it infers object geometry from the model's learned distribution resulting in high \textbf{MD} and \textbf{WE} metrics compared to our work.

\subsection{\cbEdit{Diffusion Self Guidance (DSG) \cite{epstein2024diffusion}}} 
\parahead{Implementation} \cbEdit{We ran the official implementation of DSG from the authors but it did not perform well for real-images as the authors provide code only for running on generated images.
We instead use the implementation of \cite{diffusion_self_guidance_github} and incorporated DDIM inversion to preserve details of the input image that improved the quality of results using Stable Diffusion V1.4 model.
The original work uses Imagen model which is not available.
We transform the shape using the transform $\mathcal{F}$ in our paper and use the shape guidance from Eqn. 9 of the DSG paper to penalize for movement which works better according to authors compared to centroid guidance.
We had to double the shape and appearance guidance from the default implementation for real images.}

\parahead{Timing Analysis} \cbEdit{This implementation uses 50 DDIM steps to perform edits and takes 50 seconds to edit.}

\parahead{Performance Analysis} 
\cbEdit{DSG often loses appearance details when the shape guidance is large or does not move the object when the appearance guidance is large.
This primarily occurs because it does not dis-entangle appearance and geometry accurately leading to improvement of appearance at the cost of movement or vice versa.
The geometric attention sharing mechanism of our work dis-entangles geometry from appearance leading to more accurate edits both qualitative and quantitatively (see manuscript Tab. 1, Fig. 4 and supplement Fig. \ref{fig:comparison})}

Note that we use prompts for baselines: FreeDrag, Dragon Diffusion, Diffusion Handles, Diffusion Self Guidance and do not require prompts for editing using \coolname.
Additionally, we perform all timing analysis using Nvidia RTX 3090 on the same node.
The metric evaluations for all the methods use the default editing parameters from the official implementation unless mentioned otherwise above.

\section{Edit Attention Progression}
\begin{figure}[H]
\centering
\vspace{-0.1in}
\scalebox{1.0}{
\centering
{\includegraphics[width=0.48\textwidth, trim={0.4cm 0 0.4cm 0.4cm}, clip]{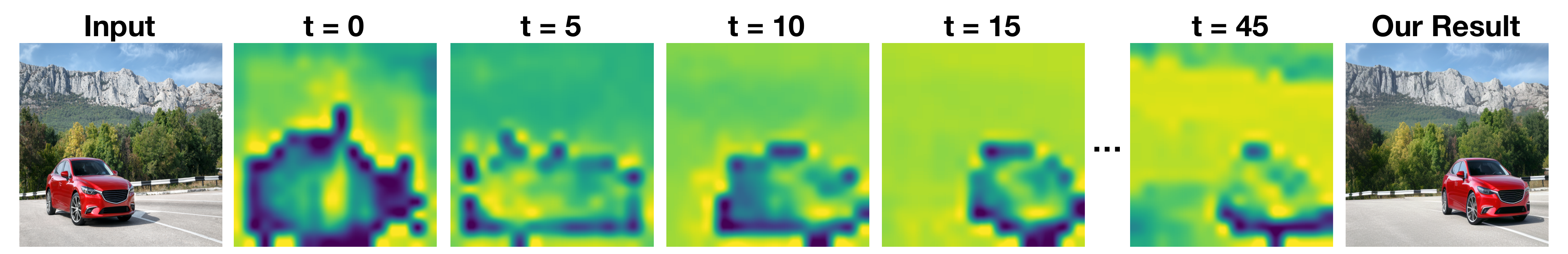}
}}
\vspace{-0.28in}
\caption{\textbf{Attention Progression}: We visualize the principal components of the self attention maps within the first \textit{up-block} layer during editing.
At earlier time steps ($t=5$), the attention is transitioning to move the car, but eventually moves the car to the desired location at $t=45$.
Transforming the attention map shifts the attention corresponding to the shadow of the car.
}
\vspace{-0.15in}
\label{fig:attention_progression}
\end{figure}

We show the edit progression over different reverse diffusion time-steps in \Cref{fig:attention_progression}. We visualize the top principal component of the self-attention map and show the movement of the car as the optimization progresses. 
Note that the shadow (dark) region in the attention map also shifts with the car.
Transforming the reference query and then computing the attention map transforms the shadows as well (see \Cref{fig:attention_progression}).

\parahead{Camera Projection} We set the camera FOV 49.92$^{\circ}$ for all edits in our work and we \underline{do not} require any dataset specific camera intrinsic matrix. 


\section{Metrics}
\parahead{Mean Distance (MD)} We use the mean distance metric from \cite{shi2023dragdiffusion}. 
\cite{shi2023dragdiffusion} perform drag based edits in their work and have source as well as their corresponding target drag locations. 
The mean distance metric computes correspondences between the input and the edited image using DiFT \cite{dift} and then estimates the difference between the target edit location and the predicted target location obtained using DiFT. 
In our case, all pixels in the object foreground become the source edit location, however, finding DiFT correspondences for each foreground pixel is very compute intensive. 
Hence, we find interest points using SiFT \cite{sift} in the foreground of the source image and treat them as the source edit location. 
We can then obtain the target edit location using the transform $\mathcal{F}$ estimated using camera projection. 
We then compute DiFT correspondences for these interest points and compute the mean distance metric. 

\parahead{Warp Error (WE)} The mean distance metric only measures edit adherence for interest points. 
We instead warp foreground of the source image and compute an L1 error.
This metric measures the error between the warped foreground source image and the edited image.
It measures preservation of the foreground object as well as how well it adheres to the edit.

The mean distance is analogous to Re-projection error and the Warp Error is analogous to Photometric error from the Computer Vision literature.

\parahead{Clip Similarity (CS)} We often notice degrade in background and content preservation after the edit. 
To ensure that the edits do not degrade the contents of the image, we compute the clip image embeddings \cite{Radford2021LearningTV_clip} of the source and the edited image. 
We then use these embeddings to estimate the cosine similarity between them to measure content preservation between them.

A good editing approach should have low Mean Distance (MD) and Warp Error (WE) as well as have high Clip Similarity (CS).

\begin{figure}[!h]
\centering
\vspace{-0.1in}
\includegraphics[width=0.49\textwidth]{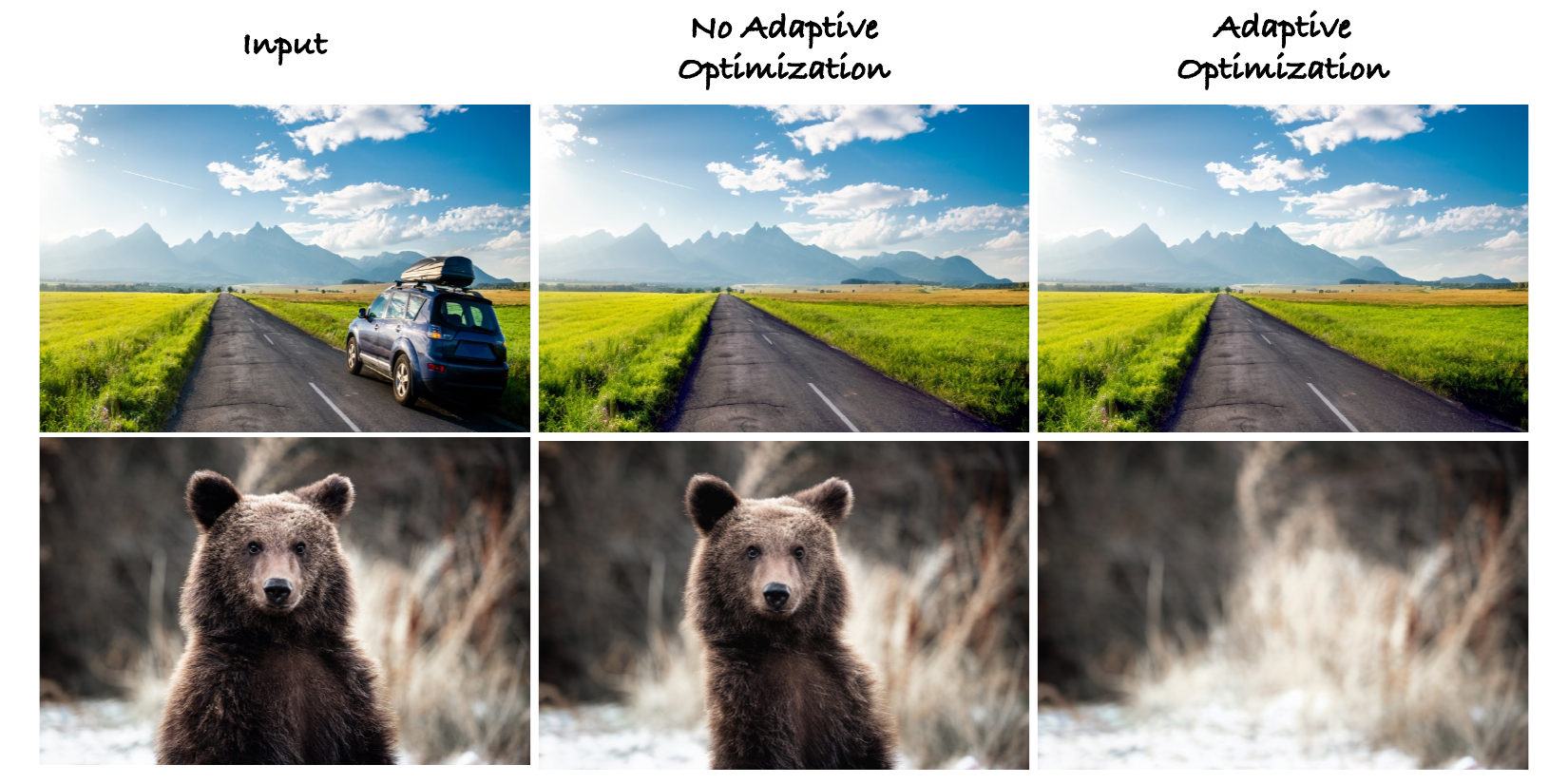}
\vspace{-0.32in}
\caption{Ablation of adaptive optimization.
Without adaptive optimization, the same losses successfully inpaint some images while others fail (middle row).
With our adaptive optimization, the same loss function works well for any image.}
\label{fig:ablation_adaptive_optim}
\vspace{-0.15in}
\end{figure}

\begin{figure}[H]
\vspace{-0.15in}
\centering
\scalebox{1.0}{
\centering
\includegraphics[width=0.48\textwidth]{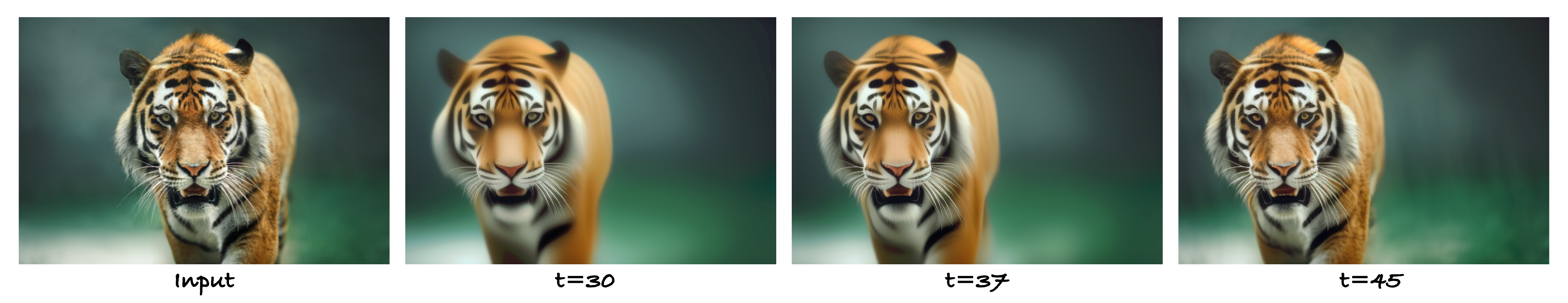}
}
\vspace{-0.3in}
\caption{\textbf{Geometry Guidance}:
Increasing steps $t$ for geometric attention sharing better preserves object style (translation edit).}
\label{fig:ablation_guidance}
\vspace{-0.2in}
\end{figure}

\section{Ablations} 
We perform a visual ablation of our design choices.
\Cref{fig:ablation_guidance,fig:ablation_adaptive_optim} shows the importance of the attention sharing mechanism and adaptive optimization.
We can see a degradation in style preservation of the edit when we don't perform geometric attention sharing till step $45$.
Without the adaptive optimization, we need image specific tuning for loss weights which is not scalable. 

\begin{figure}[h]
\centering
\vspace{-0.15in}
\includegraphics[width=0.48\textwidth]{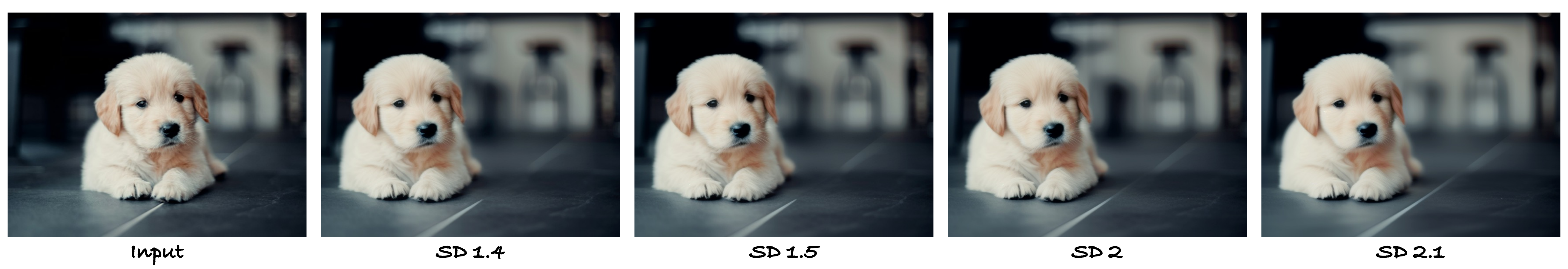}
\vspace{-0.25in}
\caption{\textbf{Editing ablation using different Stable Diffusion Models}:
We perform the same edit using different versions of Stable Diffusion.
Notice how the line is incomplete in some cases and the inpainted backgrounds are different. Our geometric attention sharing mechanism ensures that the foreground adheres to the edit and stays the same.
}
\vspace{-0.3in}
\label{fig:ablation_SD_models}
\end{figure}

In \Cref{fig:ablation_SD_models}, we use our general editing framework to perform the same edit using various Stable Diffusion models. 

\section{Perceptual Study}

We conducted a perceptual study with 53 participants to measure the efficacy of inpainting the background and benchmark \coolname against Zero123-XL.
Our perceptual study was conducted using Qualtrics \cite{qualtrics}. 
We first conducted a pilot study having 2 images per division type with 3 users to ensure that all questions are clear. 
These users did not participate in the final study. 
After getting feedback from the pilot study we conducted the full study. 
Each participant completed the study within 10 minutes. 
They were allowed to click and enlarge images for better inspection. 
We randomized the order of options presented in the study to avoid biases.
In total we presented 70 images (30 for removal, 40 for other transforms) from our dataset.
The questions were divided in three categories: edit realism (ER), edit adherence (EA), and removal edit realism (RRE).
\begin{figure}[ht]
    \vspace{-0.1in}
    \centering    \includegraphics[width=0.45\textwidth, trim={0cm 11cm 0cm 0cm}, clip]{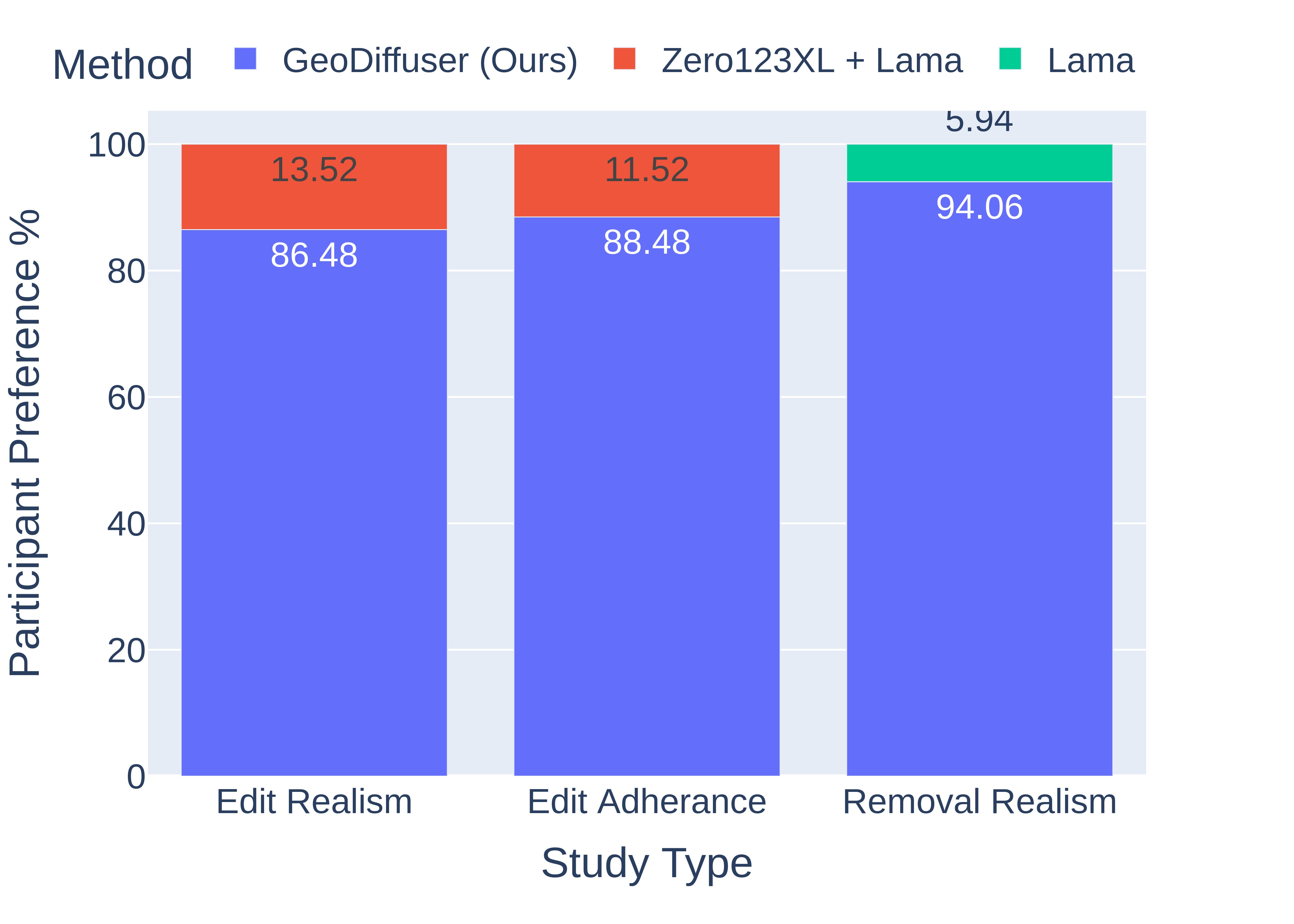}
    \vspace{-0.1in}
    \captionof{figure}{Results from perceptual study show that participants prefer our edits over \cite{liu2023one} and \cite{lama-inpainting} in a majority of the cases. \label{fig:user_study_graph}}
    \vspace{-0.1in}
\end{figure}

For removal, we generated results with LaMa~\cite{lama-inpainting}, and for the remaining two categories, results were generated with Lama followed by Zero123-XL~\cite{liu2023zero}.
Each participant answered 12 ER questions, 12 EA questions and 6 RRE questions, for a total of 30 visual questions.
In total 53 users participated in the study for which they received no compensation.

\Cref{fig:user_study_graph} shows the participant preference rate for each division of the study. For RRE, out of the 318 choices, participants preferred our method in 94.06\% of the time, which shows that \coolname is better able to inpaint the disoccluded background regions, especially removing shadows (see \Cref{fig:gallery_sup_1}).

\begin{algorithm*}
\caption{Object Removal Loss Algorithm}\label{alg:removal}
\begin{algorithmic}
\Require ${}^{r}Q, {}^{r}K, {}^{e}Q, {}^{e}K$
\Ensure $\mathcal{L}_{remove} := \text{RemovalLoss}({}^{r}Q, {}^{r}K, {}^{e}Q, {}^{e}K)$
\If{SelfAttentionBlock}
    \State ${}^{\mathcal{G}}A_{edit} := \text{AM}({}^{e}Q, {}^{r}K)$ \Comment{Shared Attention Map}
\ElsIf{CrossAttentionBlock}
    \State ${}^{\mathcal{G}}A_{edit} := \text{AM}({}^{e}Q, {}^{e}K)$ \Comment{Shared Attention Map}
\EndIf
\State ${}^{\mathcal{G}}A_{ref} := \text{AM}({}^{r}Q, {}^{r}K)$
\State $\rho_{obj \rightarrow bg}, u_{bg} := \text{torch\_max}(\text{torch\_bmm}({}^{\mathcal{G}}A_{edit}, {}^{\mathcal{G}}A_{ref}) \odot M_{bg}, -1)$ \Comment{Foreground to background correlation}
\State $\rho_{obj \rightarrow obj}, \_ := \text{torch\_max}(\text{torch\_bmm}({}^{\mathcal{G}}A_{edit}, {}^{\mathcal{G}}A_{ref}) \odot M_{obj}, -1)$\Comment{Foreground to foreground correlation}
\State $d_{obj \rightarrow bg} := \text{NormalizedCoordinateDistance}(u_{bg})$ \Comment{Coordinate distance to the background location having maximum correlation}
\State $\mathcal{L}_{remove} := \text{mean} \left( e^{-d_{\text{obj}\rightarrow \text{bg}}} (ln(\rho_{\text{obj}\rightarrow \text{obj}}) - ln(\rho_{\text{obj}\rightarrow \text{bg}}))\right)$
\end{algorithmic}
\end{algorithm*}

For ER, our method was preferred 86.48\% out of 636 cases.
This demonstrates that \coolname preserves object style better than other methods, especially in transforming shadows and reflections.
Finally, for EA we included included $16$ 2D and $24$ 3D edits.
Our method was preferred 88.48\% out of 636 cases. This demonstrates that our method more faithfully performs the intended edit, even challenging ones such as 3D rotation. 
Whereas the baseline is only capable of performing edits from a more narrow range.



\section{Failure Cases}
\begin{figure}[ht]
\centering
\vspace{-0.1in}
\includegraphics[width=0.48\textwidth]{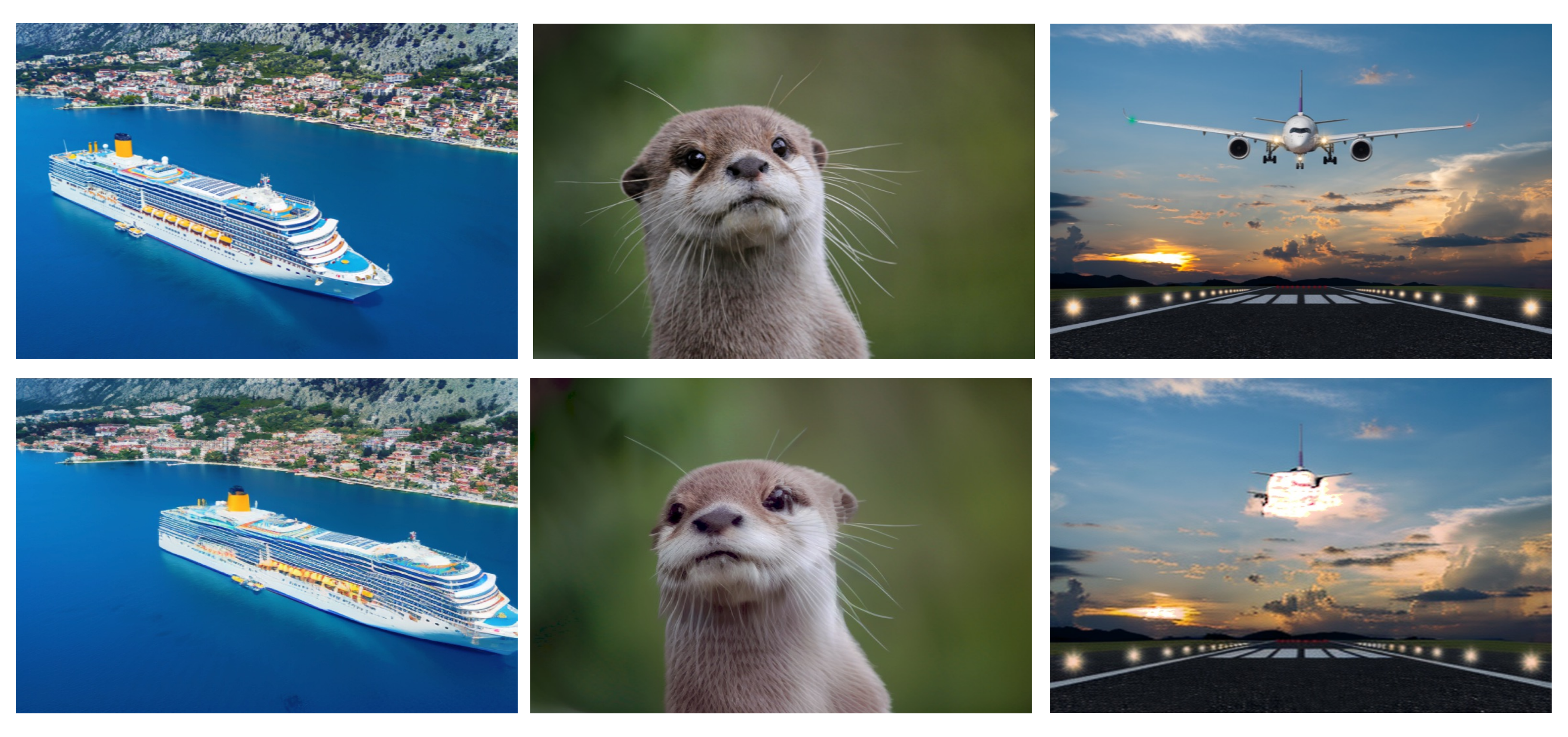}
\vspace{-0.3in}
\caption{\textbf{Failure Cases}: Each example presents the input image at the top followed by the edited image at the bottom. 
As our geometric edits are performed in a lower dimensional latent space, we face aliasing and interpolation artefacts as shown in the yellow regions of the ship (left). 
Occasionally our optimization results in sub-optimal solutions for foreground (middle) and background dis-occlusions (right). 
}
\vspace{-0.3in}
\label{fig:failure_cases}
\end{figure}

\Cref{fig:failure_cases} displays examples where our method does not perform well.
The generation capabilities of the diffusion model at times produces sub-optimal solutions for foreground and background of the image.
Additionally, similar to prior works, we can not generate novel views with large rotations and this is a future direction to explore.

\section{Miscellaneous Edits}
\begin{figure}[h]
\centering
\vspace{-0.15in}
\includegraphics[width=0.48\textwidth]{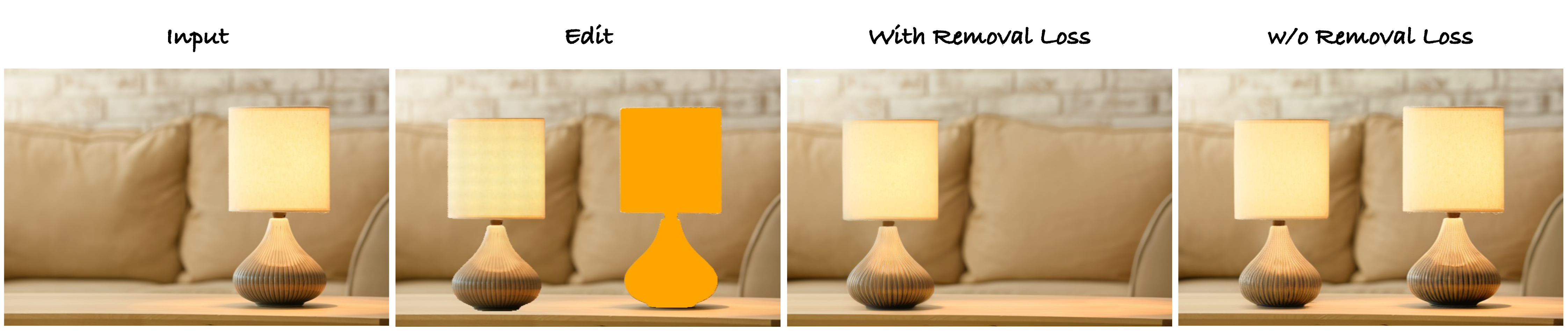}
\vspace{-0.25in}
\caption{Foreground duplication by reducing the turning off optimization or setting the removal loss weight to zero.
}
\vspace{-0.3in}
\label{fig:duplication}
\end{figure}

Our method enables object duplication by turning off the optimization or setting the removal loss to 0 (\Cref{fig:duplication}).

\section{Diffusion Correction}
\label{sec:supp_transform_correction}
\begin{figure}[h]
\centering
\vspace{-0.15in}
\includegraphics[width=0.48\textwidth]{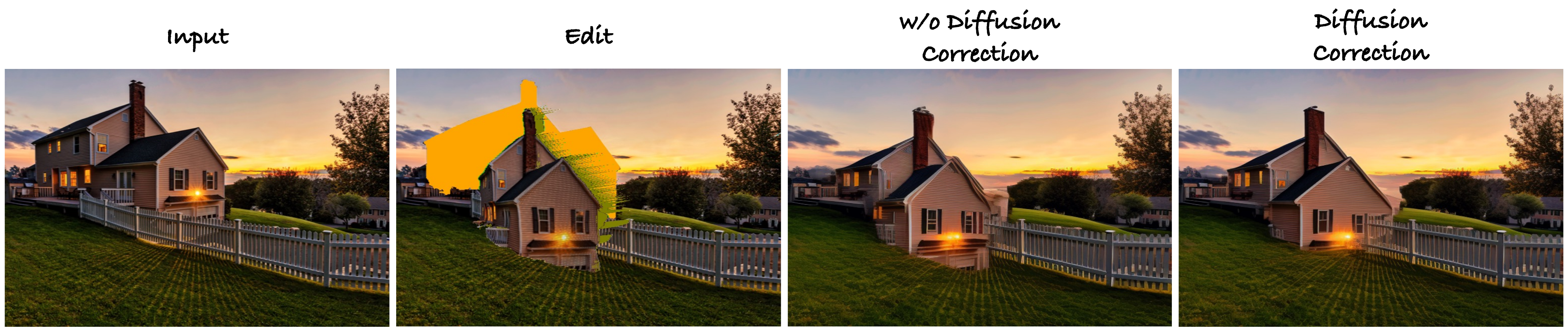}
\vspace{-0.25in}
\caption{Diffusion Correction to correct transforms and aliasing.
}
\vspace{-0.3in}
\label{fig:diffusion_correction}
\end{figure}

\begin{figure*}[t]
\centering
\includegraphics[width=0.98\textwidth]{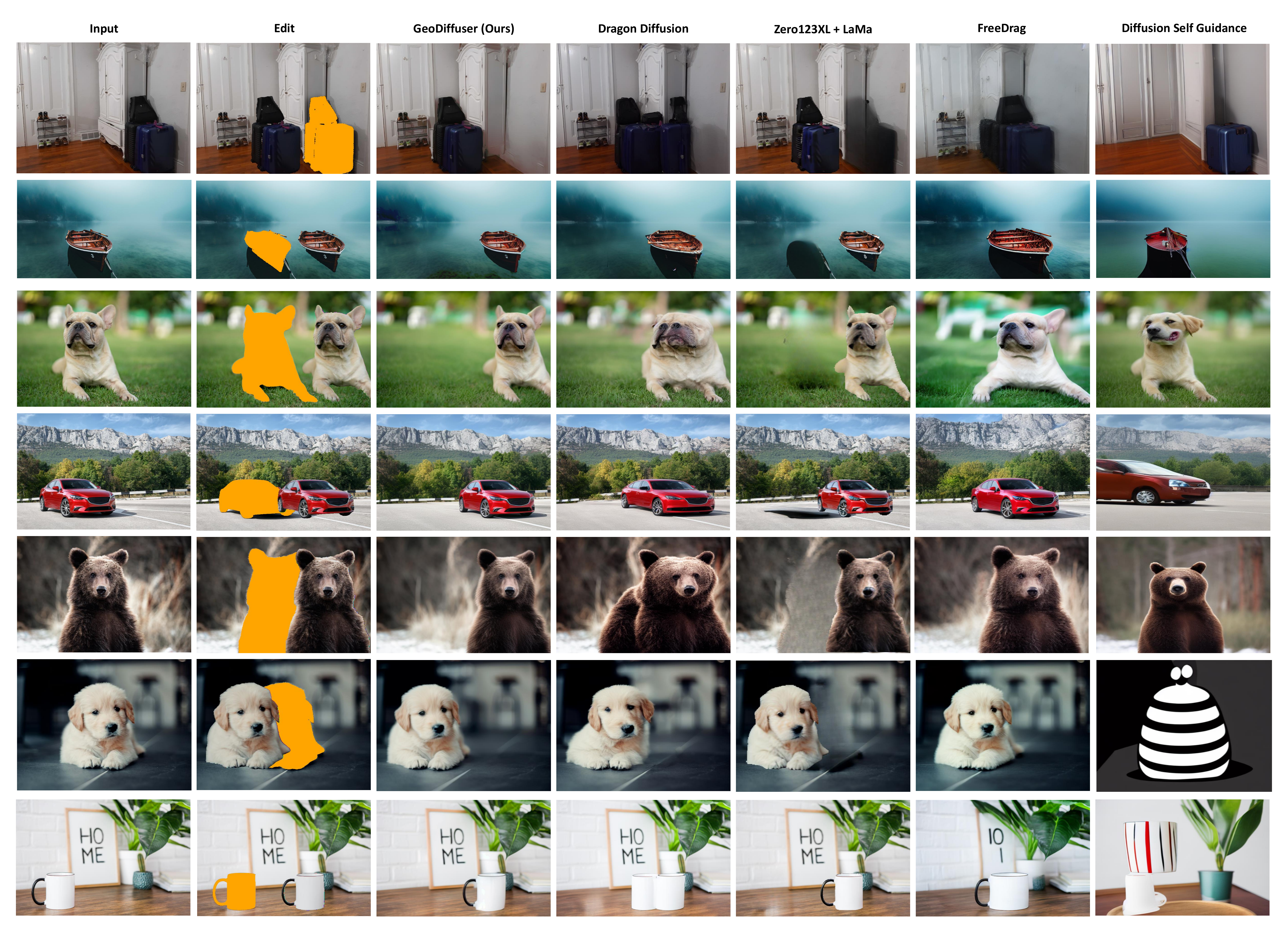}
\vspace{-0.2in}
\caption{We perform the same edit using prior works and compare with out work. We show 2D edits here as Dragon Diffusion can not perform 3D edits. 
We show the intended 2D edit in column 2 where the orange mask determines the region to be inpainted and the green regions determine the region to be filled with foreground. 
Note that Dragon Diffusion \cite{mou2023dragondiffusion} \& FreeDrag \cite{ling2023freedrag} requires prompts along with the edit and our method does not. 
FreeDrag does not remove the object from the source location appropriately resulting in stretching it.
}
\vspace{-0.2in}
\label{fig:comparison}
\end{figure*}

\begin{figure*}[!ht]
\centering
\scalebox{0.98}{
\centering
\includegraphics[width=0.98\textwidth]{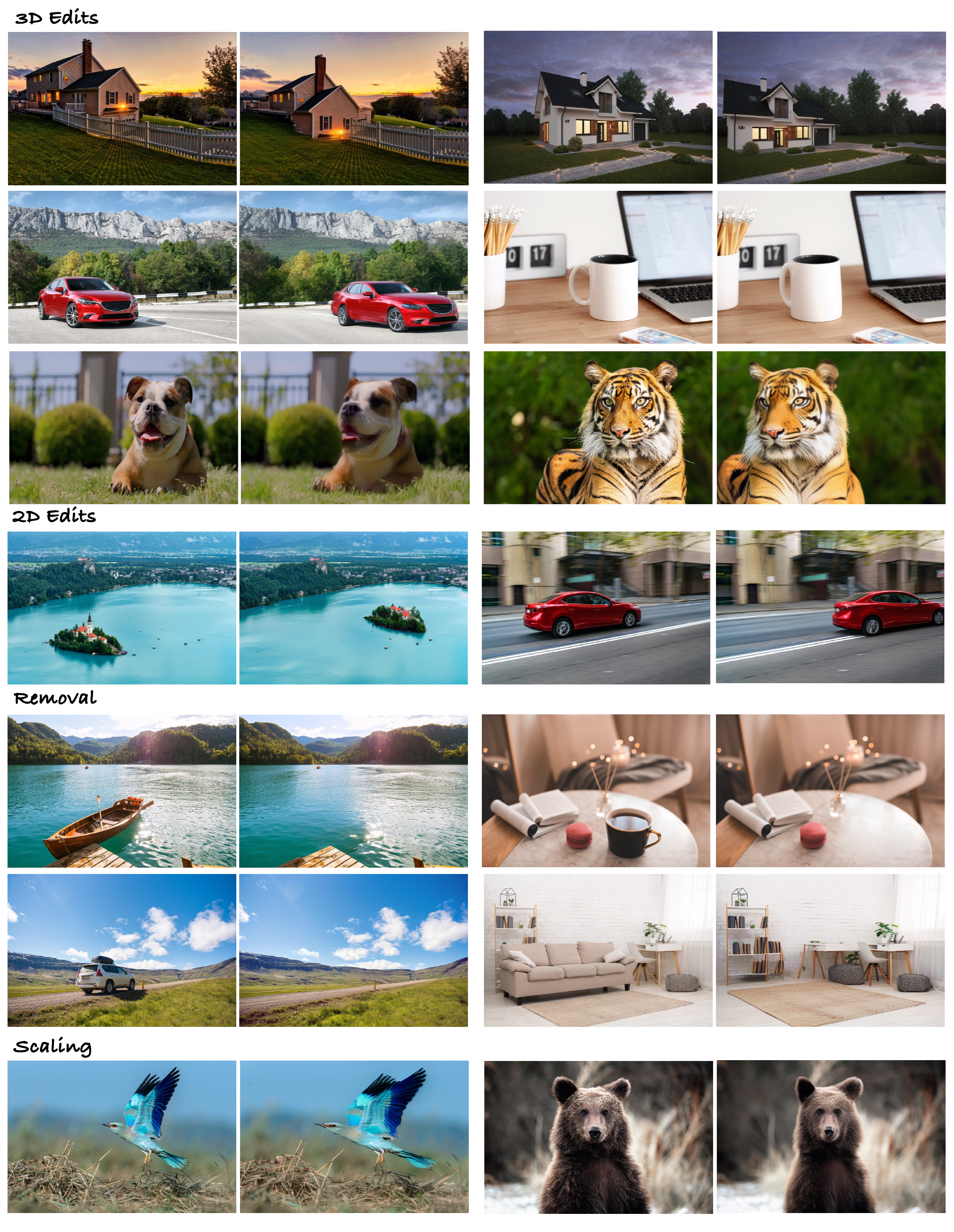}
}
\vspace{-0.25in}
\caption{Qualitative results showing all variations of 2D and 3D edits performed by \textbf{\coolname} on natural images.
Notice how our method not only removes/transforms objects but also the object's reflection and shadows (car, couch, boat). 
For 3D edits, our method produces plausible results for rotations as high as 30$^\circ$.
For scaling, we can perform both uniform and non-uniform scaling operations.
}
\label{fig:gallery}
\end{figure*}

\begin{figure*}[ht]
\centering
\vspace{-0.3in}
\makebox[\textwidth][c]{\includegraphics[width=1.0\textwidth, trim={0cm 2cm 0 0}, clip]{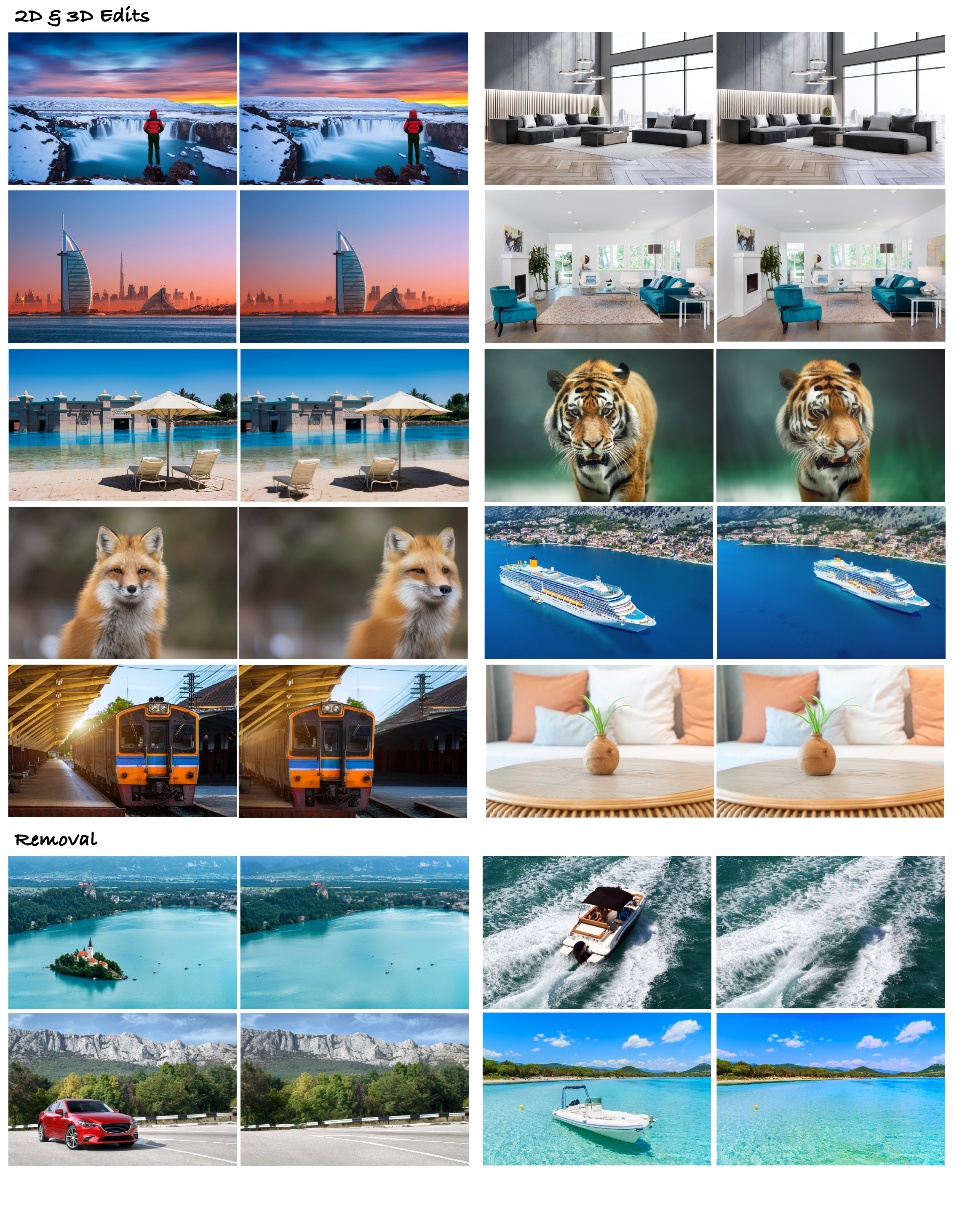}
  \vspace{-1.0in}}
\vspace{-0.35in}
\caption{
We display more qualitative results of our method. 
Each example has the input image in the left and the result of the edit in the right.
}
\label{fig:gallery_sup_1}
\end{figure*}

Occasionally, edit transforms $\mathcal{F}$ are incorrect. 
For instance, a straight line might be mapped to a jagged curved line.
In these cases, it is important for the editing method to marginally disregard the desired edit and preserve the content of the image.
This reduces adherence to the edit and produces better results. 
We can also control this in our attention sharing mechanism by allowing the diffusion model to self-correct and find correspondences for more realistic results as shown in \Cref{algo:attention_sharing}. 
This plays a crucial role in edits with sharp geometric structures such as buildings etc (see \Cref{fig:diffusion_correction}). 
We enable Diffusion Correction for the last 15 reverse diffusion steps in our experiments.

\section{Object Removal}
We detail the object removal loss in \Cref{alg:removal}.

\section{Amodal Loss}
Transforming foreground objects drastically introduces depth smearing. 
We add a small penalty to each edit to force inpainting of the foreground object in these smeared regions using the amodal loss on the amodal mask $M_{amodal}$ obtained by interpolating features after reprojection as
\begin{align}
    \mathcal{L}_{amodal} := \text{mean}(M_{amodal} \cdot || {}^{\mathcal{G}}Y_{\text{edit}} - \text{interp}({}^{\mathcal{G}}Y_{\text{ref}}) ||_1).
\end{align}

\section{Future Work \& Impact}
We present \coolname, a method that performs geometric transform on objects to edit real-world images. 
Our method only requires performing geometric manipulation to the attention layers of the model along with optimization to perform the desired edit.
This assumption makes our method very general and better adhere to edits that can be leveraged by future works for geometric analysis of diffusion models and editing in video diffusion models.
Another interesting future direction is to perform unsupervised novel view synthesis for real-world scenes by leveraging key ideas from our work that might be able to improve Score Distillation Sampling \cite{poole2023dreamfusion}.

\section{\cbEdit{Discussion on Concurrent Works that Train on Video Data}}
\cbEdit{Concurrent works such as InstaDrag~\cite{Shi2024InstaDragLF}, DragNUWA~\cite{Yin2023DragNUWAFC}, \& MagicFixup~\cite{Alzayer2024MagicFS} perform drag edits by training over video data. 
We detail the advantages \& disadvantages of these works and similar works without testing some of these implementations as they are not public.
Two advantages of these works include: 1) the inpainting for in/near-distribution images will be accurate with better novel view synthesis of foreground object and 2) faster inference. 
However, these methods and in-general video diffusion models have the following dis-advantages that need further exploration: 1) They require large scale training datasets and heavy compute for training \& do not leverage the capabilities of existing diffusion models as in our work. 2) moving foreground most often introduces background movement as video datasets do not distinguish between foreground and background motion, 3) these methods do not bake geometry within their architecture leading to edits that may not be 3D consistent, 4) they are trained with optical flow within a bounded range and often lose object details and identity when the desired edit motion is beyond this range, and lastly 5) they do not explore having inference time optimization disabling the user to control different aspects of the edit by merely changing loss weights. 
We believe that the geometry attention sharing mechanism and loss functions from \coolname can help improve these models to ensure edits and generation that are consistent with geometry in future works.}

\section{Discussion on Slider based UI as opposed to Drag UI}
We follow the slider UI of zero123 \cite{liu2023zero}. 
It is easy to control precise rotations as well as preserve the geometry using sliders as compared to a drag-based UI. 
However, we can also have a drag-based UI if the user prefers, however, this makes controlling rotations difficult.

    




\section{User Interface}
\begin{figure*}[!ht]
\centering
\scalebox{0.9}{
\includegraphics[width=0.95\textwidth, trim={0cm 74cm 0 0}, clip]{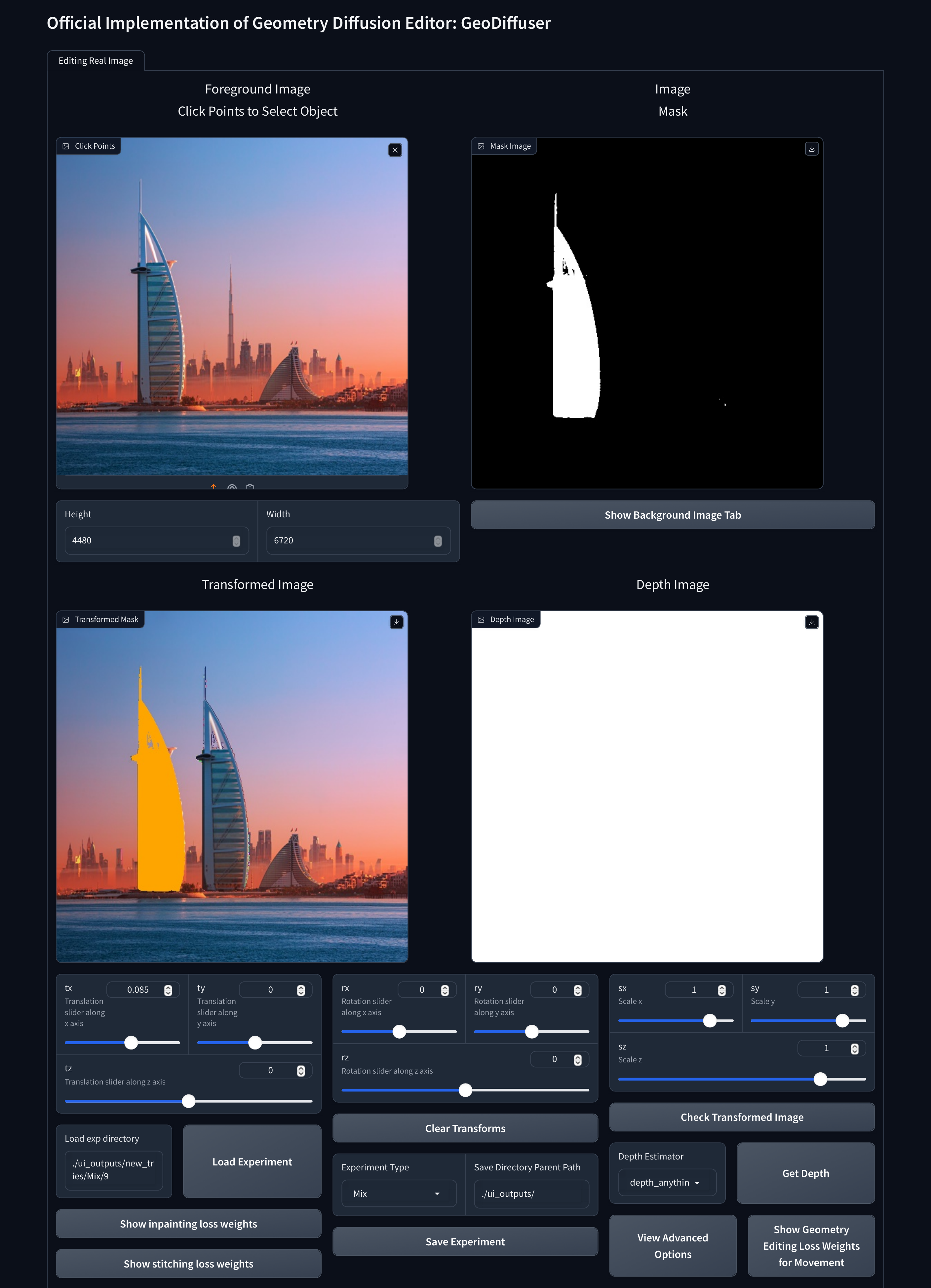}
}
\caption{\coolname UI that allows users to edit images in the wild. We provide options for users to choose a monocular depth model for geometric editing. The transformed image represents the edit that the user wishes to perform. Here, the orange mask displays the region that needs to be inpainted.}
\label{fig:ui}
\end{figure*}

See \Cref{fig:ui,fig:ui_2} that display the user interface used to perform edits using \coolname. 
We develop this user interface using Gradio \cite{gradio}. 
We also submit a video along with this supplement that displays the editing process performed by a user and a website that shows gifs of edits using \coolname.

\begin{figure*}[!ht]
\centering
\scalebox{0.85}{
\includegraphics[width=0.9\textwidth, trim={0cm 0cm 0 66cm}, clip]{images/UI/UI_demo_2_small.pdf}
}
\caption{\coolname UI also provides options for varying parameters for editing. The edited image in the bottom displays the image after the edit is complete.\label{fig:ui_2}}

\end{figure*}

\section{\cbEdit{Complex Shapes and Human Edits}}

\begin{figure*}[h]
\centering
\includegraphics[width=0.96\textwidth]{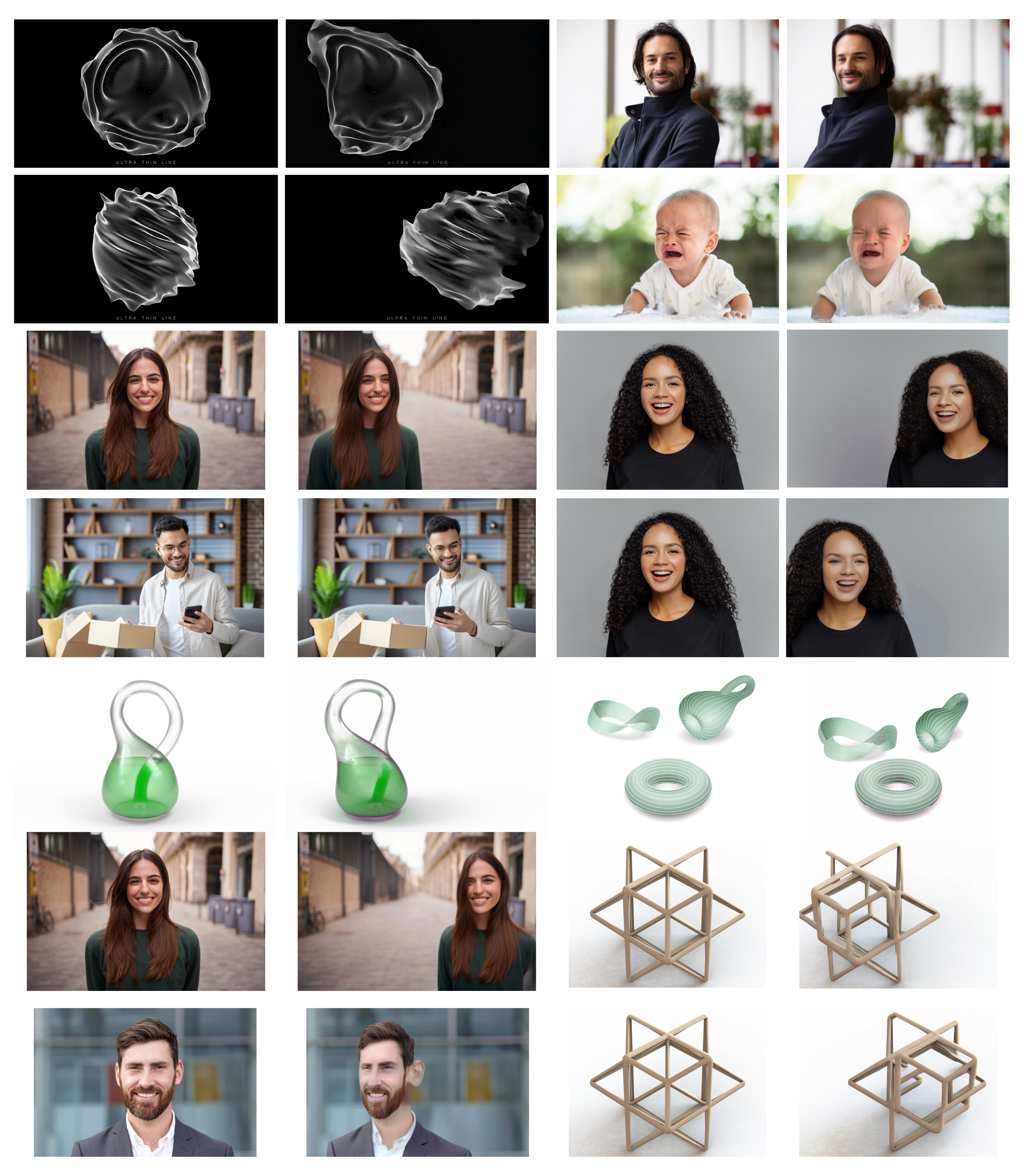}
\vspace{-0.25in}
\caption{\cbEdit{Editing Complex Geometries and Humans. 
For each row, the left shows the input image and the right shows the result of the edit.
Our method provides plausible edits for most cases of complex 3D shapes and humans even when the model has not seen this.
\textbf{Last row} shows some limitations of our work where the ear is interpolated because of editing at low resolution and smearing in depth maps.
Our edits are limited by the base model wherein there are some cases where the face/complex shape loses detail because the model has not seen these during training. 
We also notice that at times the model opens eyes even when the eyes are closed in the input image because of training bias in the stable diffusion base model.
}
}
\label{fig:complex_shapes}
\end{figure*}

\cbEdit{Our method generates plausible edits for complex 3D shapes and close-up humans images (\Cref{fig:complex_shapes}). 
However, our method finds it challenging to preserve arms and legs in far shots of humans.}

\vfill
\hfill
\newpage
{\small
\bibliographystyle{ieee_fullname}
\bibliography{egbib}
}

\end{document}